\documentclass{article}

\usepackage{microtype}
\usepackage{graphicx}
\usepackage[aboveskip=2pt]{subcaption}
\usepackage{booktabs} 
\usepackage{nicefrac}
\usepackage{amsmath,amsfonts,amssymb,mathrsfs}

\usepackage{hyperref}
\usepackage[numbers,square,sort&compress]{natbib}
\usepackage[capitalize,nameinlink]{cleveref}
\usepackage{xcolor}
\usepackage{url}
\usepackage{colortbl}

\usepackage{tcolorbox}

\crefname{section}{Sec.}{Sects.}
\crefname{proposition}{Prop.}{Props.}
\crefname{lemma}{Lem.}{Lems.}
\crefname{model}{Mod.}{Mods.}
\crefname{appendix}{App.}{Apps.}

\usepackage[accepted]{icml2020}

\newlength\figureheight
\newlength\figurewidth

\usepackage{bm}
\newcommand{\mathbold}[1]{\bm{#1}}
\newcommand{\mbf}[1]{\mathbf{#1}}
\newcommand{\vect}[1]{\mbf{#1}}

\usepackage{xspace}
\newcommand{\eg}{\textit{e.g.}\xspace}
\newcommand{\ie}{\textit{i.e.}\xspace}

\newcommand{\etc}{\textit{etc.}\xspace}

\newcommand{\T}{\top}    
\newcommand{\dd}{\,\mathrm{d}} 
\newcommand{\E}{\mathbb{E}}    
\newcommand{\R}{\mathbb{R}}    
\newcommand{\N}{\mathrm{N}}   

\DeclareMathOperator{\Cov}{Cov}
\DeclareMathOperator{\Var}{Var}

\newcommand{\kron}{\raisebox{1pt}{\ensuremath{\:\otimes\:}}} 

\newcommand{\diff}[2]{\mathrm{\frac{\partial\mathit{#1}}{\partial\mathit{#2}}}}

\newcommand{\diffIIvec}[3]{\mathrm{\frac{\partial^2\mathit{#1}}{\partial\mathit{#2}\partial\mathit{#3}}}}

\newcommand{\vlambda}[0]{\mathbold{\lambda}}

\newcommand{\vmu}[0]{\mathbold{\mu}}
\newcommand{\vsigma}[0]{\mathbold{\sigma}}

\newcommand{\vtheta}[0]{\mathbold{\theta}}

\newcommand{\MSigma}[0]{\mathbold{\Sigma}}

\newcommand{\MOmega}[0]{\mathbold{\Omega}}

\renewcommand{\mid}[0]{\,|\,}

\newcommand{\GP}{\mathcal{GP}}

\newcommand{\vb}{\mbf{b}}

\newcommand{\ve}{\mbf{e}}
\newcommand{\vf}{\mbf{f}}
\newcommand{\vg}{\mbf{g}}
\newcommand{\vh}{\mbf{h}}

\newcommand{\vm}{\mbf{m}}

\newcommand{\vq}{\mbf{q}}
\newcommand{\vr}{\mbf{r}}

\newcommand{\vv}{\mbf{v}}
\newcommand{\vw}{\mbf{w}}
\newcommand{\vx}{\mbf{x}}
\newcommand{\vy}{\mbf{y}}

\newcommand{\MA}{\mbf{A}}
\newcommand{\MB}{\mbf{B}}
\newcommand{\MC}{\mbf{C}}

\newcommand{\MF}{\mbf{F}}
\newcommand{\MG}{\mbf{G}}
\newcommand{\MH}{\mbf{H}}
\newcommand{\MI}{\mbf{I}}
\newcommand{\MJ}{\mbf{J}}
\newcommand{\MK}{\mbf{K}}
\newcommand{\ML}{\mbf{L}}

\newcommand{\MP}{\mbf{P}}
\newcommand{\MQ}{\mbf{Q}}
\newcommand{\MR}{\mbf{R}}
\newcommand{\MS}{\mbf{S}}

\newcommand{\MW}{\mbf{W}}

\newcommand{\ME}{\mbf{E}}

\definecolor{cgray}{gray}{0.4}
\newcommand{\comm}[1]{\hfill\textcolor{cgray}{#1}}

\renewcommand{\algorithmiccomment}[1]{#1}

\renewcommand{\paragraph}[1]{{\bf #1}~~}

\definecolor{mycolor0}{rgb}{0.2667,0.4471,0.7098}
\definecolor{mycolor1}{rgb}{0.1647,0.6706,0.3804}
\definecolor{mycolor2}{rgb}{0.8275,0.2627,0.3059}
\definecolor{mycolor3}{rgb}{0.5216,0.4392,0.7176}
\definecolor{mycolor4}{rgb}{0.8118,0.7255,0.4118}
\definecolor{mycolor5}{rgb}{0.2745,0.7176,0.8157}

\definecolor{mylcolor0}{rgb}{0.6902,0.7686,0.8863}
\definecolor{mylcolor1}{rgb}{0.5451,0.8902,0.6941}
\definecolor{mylcolor2}{rgb}{0.9412,0.7490,0.7647}
\definecolor{mylcolor3}{rgb}{0.8627,0.8392,0.9176}
\definecolor{mylcolor4}{rgb}{0.9569,0.9373,0.8667}
\definecolor{mylcolor5}{rgb}{0.7529,0.9020,0.9373}
\definecolor{mylcolor6}{rgb}{0.8750,0.8750,0.8750}
\definecolor{mygrey}{rgb}{0.93, 0.93, 0.93}

\definecolor{color2}{rgb}{0.75,0.75,0}
\definecolor{color1}{rgb}{1,0.498039215686275,0.0549019607843137}
\definecolor{color0}{rgb}{0.12156862745098,0.466666666666667,0.705882352941177}

\usepackage{tabularx}
\usepackage{array}
\newcommand{\PreserveBackslash}[1]{\let\temp=\\#1\let\\=\temp}
\newcolumntype{C}[1]{>{\PreserveBackslash\centering}p{#1}}

\usepackage{tikz,pgfplots}
\usetikzlibrary{plotmarks,shapes,arrows}
\pgfplotsset{compat=newest} 

\pgfplotsset{/pgf/number format/.cd, 1000 sep={}}
\pgfkeys{/pgf/number format/.cd,1000 sep={}}

\pgfplotsset{every axis/.append style={
		grid style={line width=0.6pt,dotted,gray}}}

\pgfplotsset{every axis/.append style={
		legend style={inner xsep=1pt, inner ysep=0.5pt, nodes={inner sep=1pt, text depth=0.1em},draw=none,fill=none}
}}

\usepgfplotslibrary{groupplots}

\pgfplotsset{ignore legend/.style={every axis legend/.code={\let\addlegendentry\relax}}}

\RequirePackage{algorithm}
\RequirePackage{algorithmic}

\usepackage{comment}

\icmltitlerunning{State Space Expectation Propagation}

\begin{document}

\twocolumn[
\icmltitle{State Space Expectation Propagation: \\ Efficient Inference Schemes for Temporal Gaussian Processes}

\icmlsetsymbol{equal}{*}

\begin{icmlauthorlist}
	\icmlauthor{William J.\ Wilkinson}{aalto}
	\icmlauthor{Paul E.\ Chang}{aalto}
	\icmlauthor{Michael Riis Andersen}{dtu}
	\icmlauthor{Arno Solin}{aalto}
\end{icmlauthorlist}

\icmlaffiliation{aalto}{Department of Computer Science, Aalto University, Finland}
\icmlaffiliation{dtu}{Department of Applied Mathematics and Computer Science, Technical University of Denmark, Denmark}

\icmlcorrespondingauthor{William J.\ Wilkinson}{william.wilkinson@aalto.fi}

\icmlkeywords{Gaussian processes, approximate inference, state space models}

\vskip 0.3in
]

\printAffiliationsAndNotice{}  

\begin{abstract}
We formulate approximate Bayesian inference in non-conjugate temporal and spatio-temporal Gaussian process models as a simple parameter update rule applied during Kalman smoothing. This viewpoint encompasses most inference schemes, including expectation propagation (EP), the classical (Extended, Unscented, \etc) Kalman smoothers, and variational inference. We provide a unifying perspective on these algorithms, showing how replacing the power EP moment matching step with linearisation recovers the classical smoothers. EP provides some benefits over the traditional methods via introduction of the so-called cavity distribution, and we combine these benefits with the computational efficiency of linearisation, providing extensive empirical analysis demonstrating the efficacy of various algorithms under this unifying framework. We provide a fast implementation of all methods in JAX.
\end{abstract}

\begin{figure}[t!]
	\centering\small
	\pgfplotsset{yticklabel style={rotate=90}, ylabel style={yshift=0pt},scale only axis,axis on top,clip=true,clip marker paths=true}
	\setlength{\figurewidth}{.45\columnwidth}
	\setlength{\figureheight}{0.96\figurewidth}
	\begin{subfigure}[b]{\columnwidth}
		\centering
		\input{./fig/banana-ek-1.tex}
		\hfill
		\input{./fig/banana-ek-2.tex}
		\vspace{-0.2em}
		\caption{EKF forward and RTS backward pass}
	\end{subfigure}
	\\[0.4em]
	\begin{subfigure}[b]{\columnwidth}
		\centering
		\input{./fig/banana-ek-3.tex}
		\hfill
		\input{./fig/banana-ek-4.tex}
		\caption{Extended Expectation Propagation (EEP)}
		\vspace{-0.8em}
	\end{subfigure}
	\caption{Filtering and smoothing in the {\em Banana} classification task. Training data  represented by coloured points, the decision boundaries by black lines, and the predictive mean for the class label by colour map \includegraphics[width=2.5em,height=0.75em]{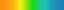}. The vertical dimension is treated as the `spatial' input and the horizontal as the sequential (`temporal') dimension. Forward sweep on the left, backward sweep on the right. Top panels (a) show the EKF; bottom (b) is the 2\textsuperscript{nd} iteration of EEP.}
	\label{fig:banana-ek}
	\vspace{-1em}
\end{figure}

\section{Introduction} \label{sec:intro}
Gaussian processes \citep[GPs,][]{rasmussen2006gaussian} are a nonlinear probabilistic modelling tool that combine well calibrated uncertainty estimates with the ability to encode prior information, and as such they are an increasingly effective method for many difficult machine learning tasks. The well known limitations of GPs are \emph{(i)}~their cubic scaling in the number of data, and \emph{(ii)}~their intractability when the observation model is non-Gaussian.

For \emph{(i)}, a wide variety of methods have been proposed \citep[\eg][]{Hensman+Fusi+Lawrence:2013,salimbeni2017doubly,Wang:2019}, with perhaps the most common being the sparse-GP approach \citep{Quinonero-Candela+Rasmussen:2005} which summarises the GP posterior through a subset of `inducing' points. However, when the data exhibits a natural ordering---as in temporal or spatio-temporal tasks---many GP priors can be rewritten in closed-form in terms of stochastic differential equations \citep[SDEs,][]{Sarkka+Solin:2019}, allowing for linear-time exact inference via Kalman filtering \citep{hartikainen2010kalman,Reece+Roberts:2010}. This link is beneficial in scenarios such as climate modelling, or audio signal analysis, which exhibit both high and low-frequency behaviour. A sparse-GP analogy to the Shannon--Nyquist theorem \citep{tobar2019band} tells us that audio signals, for example, necessarily require tens of thousands of inducing points per second of data, rendering sparse approximations infeasible for all but the shortest of time series. This strongly motivates our reformulation of temporal GPs as SDEs for efficient inference.

For limitation \emph{(ii)}, a wide variety of approximative inference methods have been considered, with the current gold-standard being various sampling schemes \citep[see][for an overview]{Gelman+Carlin+Stern+Dunson+Vehtari+Rubin:2013}, variational methods \citep[][]{Wainwright-Jordan:2008,Opper+Archambeau:2009,Titsias:2009}, and expectation propagation \citep[EP,][]{minka2001expectation,jylanki2011robust, Kuss2006,Bui:2017}. Despite Minka's original work having its foundations in filtering and smoothing, all the special characteristics of temporal models have not been thoroughly leveraged in the machine learning community. We extend recent work on approximate inference under the state space paradigm, and provide a framework that unifies EP and traditional methods such as the Extended \citep[EKF,][]{Bar-Shalom+Li+Kirubarajan:2001} and Unscented Kalman filters \citep[UKF,][]{Julier+Uhlmann+Durrant-Whyte:1995,Julier+Uhlmann+Durrant-Whyte:2000}. Our framework provides ways to trade off accuracy and computation, and we show that an iterated version of the EKF with EP-style updates can be efficient and easy to implement, whilst providing good performance in cases where the likelihood model is not locally highly nonlinear. For completeness, we also formulate variational inference in the same setting.

We show that such tools are not limited to one-dimensional-input models; instead they only require us to treat a single dimension of the data sequentially (regardless of whether it is actually ordered, or represents time). We apply our methods to multi-dimensional problems such as 2D classification (see \cref{fig:banana-ek}) and 2D log-Gaussian Cox processes.

Our main contributions are:
{\em (i)}~We formulate EP as a Kalman smoother, showing how it unifies many classical smoothing methods, providing an efficient framework for inference in temporal GPs.
{\em (ii)}~We show how to rewrite common machine learning tasks (likelihoods) into canonical state space form, and provide extensive analysis demonstrating performance in many modelling scenarios.
{\em (iii)}~We show how the state space framework can be extended beyond the one-dimensional case, applying it to multidimensional classification and regression tasks, where we still enjoy linear-time inference over the sequential dimension.
{\em (iv)}~We provide fast JAX code for inference and learning with all the methods described in this paper, available at \url{https://github.com/AaltoML/kalman-jax}.

\section{Background}
Gaussian processes \citep[GPs,][]{rasmussen2006gaussian} form a non-parametric family of probability distributions on function spaces, and are completely characterized by a mean function $\mu(t): \mathbb{R} \rightarrow \mathbb{R}$ and a covariance function $\kappa(t, t'): \mathbb{R} \times \mathbb{R} \to \mathbb{R}$. Let $\{(t_k, y_k)\}_{k=1}^n$ denote a set of $n$ input--output pairs for a time series (we first consider the 1D input case), then GP models typically take the form
\begin{equation}
    \hspace*{-1em}
	f(t) \sim \GP(\mu(t), \kappa(t,t')), \quad 
	\vy \mid \vf \sim \prod_{k=1}^{n} p(y_{k} \mid f(t_{k})), \label{eq:GP-prior-lik}
\end{equation}
which defines the prior for the latent function $f: \R \to \R$ and the observation model for $y_k$. For Gaussian observation models, the posterior distribution $p(\vf\mid\vy)$ is also Gaussian and can be obtained analytically, but non-Gaussian likelihoods render the posterior intractable and approximate inference methods must be applied.

\subsection{State Space Models for Gaussian Processes}
In signal processing, the canonical (discrete-time) state space model formulation is \citep[\eg,][]{Bar-Shalom+Li+Kirubarajan:2001}:
\begin{subequations}
\begin{align}
  \vx_k &= \vg(\vx_{k-1},\vq_{k}), \label{eq:nonlin-dyn} \\
  \vy_k &= \vh(\vx_k,\vsigma_k), \label{eq:nonlin-measurement}
\end{align}
\end{subequations}
for time instances $t_k$, where $\vx_k \in \R^s$ is the discrete-time state sequence, $\vy_k \in \R^d$ is a measurement sequence, $\vq_k \sim \N(\vect{0},\MQ_k)$ is the process noise, and $\vsigma_k \sim \N(\vect{0},\MSigma_k)$ is Gaussian measurement noise. The model dynamics (prior) are defined by the nonlinear mapping $\vg(\cdot,\cdot)$, while the observation/measurement model (likelihood) is given in terms of the mapping $\vh(\cdot,\cdot)$. We restrict the model dynamics $\vg(\cdot,\cdot)$ to be linear-Gaussian---defining an $s$-dimensional Gaussian process. The dynamical model \cref{eq:nonlin-dyn} becomes,
\begin{equation} \label{eq:state-space}
  \vx_k = \MA_{k} \vx_{k-1} + \vq_{k}, \quad \vq_{k} \sim \N(\vect{0},\MQ_k),
\end{equation}
which is characterised by the transition matrix $\MA_{k}$ and process noise covariance $\MQ_k$. Whilst we restrict our interest to latent Gaussian dynamics, the inference methods presented later apply to more general nonlinear state estimation settings, where the prior is not necessarily a GP (see \cref{sec:discussion}).

The motivation for linking the machine learning GP formalism with state space models comes from the special structure in temporal or spatio-temporal problems, where the data points have a natural ordering with respect to the temporal dimension. If the GP prior in \cref{eq:GP-prior-lik} admits a Markovian structure, the model can be rewritten in the form of \cref{eq:state-space}. We leverage the link between the kernel and state space forms of GPs \citep{Sarkka+Solin+Hartikainen:2013,Sarkka+Solin:2019}, which comes through linear time-invariant SDEs:
\begin{equation}\label{eq:sde}
  \dot{\vx}(t)=\MF\,\vx(t)+\ML\,\vw(t), ~~ \text{such that} ~~ \vf(t) = \MH \vx(t),
\end{equation}
where $\vw(t)$ is a white noise process, and $\MF\in\R^{s\times s}$, $\ML\in\R^{s\times v}$, $\MH\in\R^{m\times s}$ are the feedback, noise effect, and measurement matrices, respectively. Many widely used covariance functions admit this form exactly or approximately (\eg, the Mat\'ern class, polynomial, noise, constant, squared-exponential, rational quadratic, periodic, and sums/products thereof).  \citet{Sarkka+Solin:2019} discuss methods for constructing the required matrices for many GP models. Key to this formulation is that linear time-invariant SDEs are guaranteed to have a closed-form discrete-time solution in the form of a linear Gaussian state space model as in \cref{eq:state-space}. We leverage this link in order to apply sequential inference schemes to temporal and spatio-temporal GP models.

\subsection{Extended and Unscented State Estimation} \label{sec:EKF-UKF}
Many nonlinear variants of the Kalman filter have been developed to deal with the measurement model in \cref{eq:nonlin-measurement} \citep[see][for an overview]{Sarkka:2013}. The most widely known are the EKF \citep[\eg][]{Bar-Shalom+Li+Kirubarajan:2001} and UKF \citep{Julier+Uhlmann+Durrant-Whyte:1995}. The EKF linearises $\vh(\vx_k, \vsigma_k)$ via a first-order Taylor series expansion, which in turn results in linear Gaussian approximations to all the required Kalman update equations. We discuss the approach in detail in \cref{sec:lin}.

The UKF is a member of a wider class of Gaussian filtering methods \citep{Ito+Xiong:2000}, which approximate the Kalman update equations via statistical linearisation rather than a Taylor expansion. Statistical linearisation is generally intractable, involving expectations that must be computed numerically (shown in \cref{app:kf_update}). Choosing the Unscented transform as the numerical integration method results in the UKF, but other sigma-point methods can also be used \citep[see, \eg,][]{Ito+Xiong:2000,Wu+Hu+Wu+Hu:2005,Wu+Hu+Wu+Hu:2006,Simandl+Dunik:2009,kokkala2016a}---\eg\ using Gauss--Hermite cubature gives the Gauss--Hermite Kalman filter.

\subsection{Expectation Propagation} \label{sec:EP}
Expectation propagation (EP) is a general framework for approximating probability distributions proposed by \citet{minka2001expectation}. EP and its extension Power-EP \citep[PEP,][]{minka2004power} have been extensively studied for Gaussian process models and shown to provide state-of-the-art results \citep{jylanki2011robust, Kuss2006,  Bui:2017}. EP approximates the target distribution $p(\vf\mid\vy)$ with an approximation $q(\vf)$ that factorises in the same way as the target,
\begin{flalign} \label{eq:fact-post}
\hspace{-0.6em} p(\vf\mid\vy) \hspace{-0.2em} \propto \hspace{-0.2em} \prod_{k=1}^n p(\vy_k\mid \vf_k) p(\vf) \approx q(\vf) \hspace{-0.15em} \propto \hspace{-0.15em} \prod_{k=1}^n q^\textrm{site}_k(\vf_k)p(\vf)
\end{flalign}
The likelihood approximations $q^\textrm{site}_k(\vf_k)\approx p(\vy_k \mid \vf_k)$ are usually referred to as \textit{sites}. For GP models, the sites are chosen to be Gaussians and hence the global approximation $q(\vf)$ is also Gaussian. The sites are updated in an iterative fashion by minimizing local Kullback--Leibler divergences between the so-called \textit{tilted distributions}, $\hat{p}_k(\vf_k) = \frac{1}{Z_k}p(\vy_k\mid \vf_k)q_k^{\text{cav}}(\vf_k)$, and its approximation using the site,
\begin{align}\label{eq:KL}
{q_k^{\textrm{site}}}^* = \mathrm{arg}\,\min_{q^\textrm{site}_k} \text{KL}\left[\hat{p}_k(\vf_k) \,\|\, q^\textrm{site}_k(\vf_k)q_k^{\text{cav}}(\vf_k)\right],
\end{align}
where $q_k^{\text{cav}}(\vf_k)$ is the \textit{cavity distribution}: $q_k^{\text{cav}}(\vf_k) \propto q(\vf_k) / q^\textrm{site}_k(\vf_k)$.
The KL-divergence in \cref{eq:KL} is minimized using \textit{moment matching} \citep{minka2001expectation}, \ie $q_k$ is chosen such that the approximation $q^\textrm{site}_k q_k^{\text{cav}}$ matches the first two moments of the tilted distribution $\hat{p}_k$. This process is iterated for all sites until convergence. Power EP is a generalization of EP, where the KL-divergence is generalized to the $\alpha$-divergence \citep{Minka:2005}. \citet{minka2004power} showed that PEP can be implemented using the EP algorithm, by raising the site terms in the tilted distributions to a power of $\alpha$.

\begin{figure*}[t!]
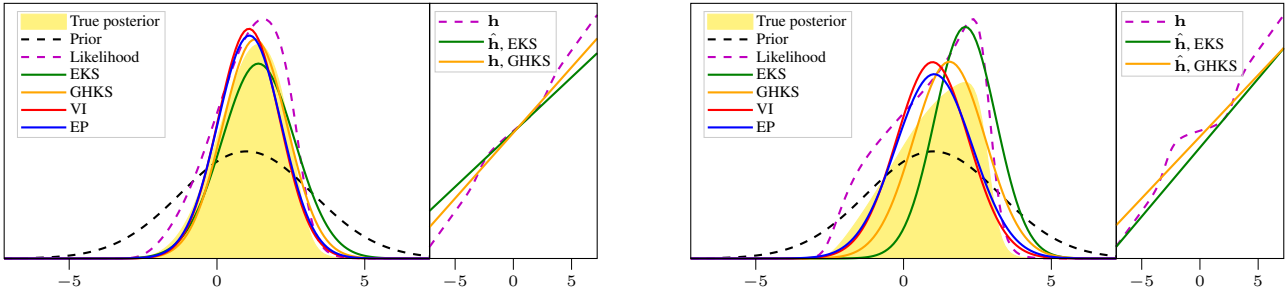

	\centering\tiny
	\pgfplotsset{yticklabel style={rotate=90}, ylabel style={yshift=0pt},scale only axis,axis on top,clip=false, ymajorticks=false}
	\setlength{\figurewidth}{.33\textwidth}
	\setlength{\figureheight}{.6\figurewidth}
	\input{./fig/approx_inf0.tex}\hspace{-0.4em}\setlength{\figurewidth}{.13\textwidth}\input{./fig/lin0.tex}
	\hfill
	\setlength{\figurewidth}{.33\textwidth}\input{./fig/approx_inf1.tex}\hspace{-0.4em}\setlength{\figurewidth}{.13\textwidth}\input{./fig/lin1.tex}
	\vspace{-1em}
	\caption{Comparison between EP, VI and iterated linearisation (EKS, GHKS). When measurement function $\vh$ is approximately linear in the region of the prior (or the cavity / posterior in the full algorithm), (\textbf{left}), linearisation $\hat{\vh}$ provides a similar result to EP / VI. When $\vh$ is highly nonlinear (\textbf{right}), the posterior approximations have different properties. 20-point Gauss--Hermite quadrature used for all methods except EKS. All methods are iterated 10 times except EP which does not require iteration for a single data point.}
	\label{fig:Linearisation}
	\vspace*{-1em}
\end{figure*}

\section{Methods}

We consider non-conjugate (\ie\ non-Gaussian likelihood) Gaussian process models with input $t$, \ie\ time, which have a dual kernel (\emph{left}) and discrete state space (\emph{right}) form for the prior \citep{Sarkka+Solin+Hartikainen:2013},
\begin{flalign} \label{eq:gp-model}
\hspace{-0.6em} \vf(t) \sim \GP\big(\vmu(t), \, \textbf{K}_{\vtheta}(t,t')\big) , \quad  \vx_k = \MA_{\vtheta,k} \vx_{k-1} + \vq_{k},
\end{flalign}
where $\vf(t) = \big( f^{(1)}(t), \hdots, f^{(m)}(t) \big)^\T \in \R^{m}$ are GPs, $\vx_k = \big( \vx_k^{(1)}, \hdots, \vx_k^{(m)} \big)^\T \in \R^{s}$ is the latent state vector containing the GP dynamics, and $\vy_k \in \R^{d}$ are observations. Each $\vx_k^{(i)}$ contains the state dynamics for one GP. Using notation $\vf_k=\vf(t_k)$, we define a time-varying linear map $\MH_k \in \R^{m \times s}$ from state space to function space, such that $\vf_k = \MH_k \vx_k$ (the time-varying mapping allows us to naturally incorporate spacial inducing points when considering multidimensional input models, see \cref{sec:s-t}). The likelihood (\emph{left}) / state observation model (\emph{right}) are
\begin{align}
\vy_{k} & \sim p(\vy_{k} \mid \vf_k) \,\,\,\quad\text{vs.}\quad\,\,\, \vy_k = \vh(\vf_k, \vsigma_k).
\end{align}
Measurement model $\vh(\vf_k, \vsigma_k)$ is a (nonlinear) function of $\vf_k$ and observation noise $\vsigma_k \sim \N(\bm{0}, \MSigma_k)$, and can generally be derived for continuous likelihoods or approximated for discrete ones by letting $\vh(\vf_k, \vsigma_k) \approx \E[\vy_k \mid \vf_k] + \bm{\varepsilon}_k$, $\bm{\varepsilon}_k\sim\N(\bm{0},\Cov[\vy_k \mid \vf_k])$. See \cref{sec:results} and \cref{sec:app-models} for derivations of some common models.
We aim to calculate the posterior over the states, $p(\vx_k \mid \vy_1, \hdots, \vy_n)$, known as the \emph{smoothing} solution, which can be obtained via application of a Gaussian filter (to obtain the \emph{filtering} solution $p(\vx_k \mid \vy_1, \hdots, \vy_k)$) followed by a Gaussian smoother. If $\vh(\cdot,\cdot)$ is linear, \ie\ $p(\vy_k \mid \vf_k)$ is Gaussian, then the Kalman filter and Rauch--Tung--Striebel \citep[RTS,][]{Rauch+Tung+Striebel:1965} smoother return the closed-form solution.

\subsection{Power EP as a Gaussian Smoother}
Our inference methods approximate the filtering distributions with Gaussians, $p(\vx_k \mid \vy_{1:k}) \approx \N(\vx_k \mid \vm_k^{\mathrm{filt}}, \MP_k^{\mathrm{filt}})$. The prediction step remains the same as in the standard Kalman filter: \mbox{$\vm_k^{\mathrm{pred}} = \MA_{\vtheta,k} \vm_{k-1}^{\mathrm{filt}}$}, and \mbox{$\MP_k^{\mathrm{pred}} = \MA_{\vtheta,k} \MP_{k-1}^{\mathrm{filt}} \MA_{\vtheta,k}^\T + \MQ_{\vtheta,k}$}. The resulting distribution provides a means by which to calculate the EP cavity, $q_k^\mathrm{cav}(\vf_k)=\N(\vf_k \mid \vmu_k^{\mathrm{cav}}, \MSigma_k^{\mathrm{cav}} )$, on the first forward pass:
\begin{equation}
  \vmu_k^{\mathrm{cav}} = \MH_k \vm_k^{\mathrm{pred}} , \quad \MSigma_k^{\mathrm{cav}} = \MH_k\MP_k^{\mathrm{pred}}\MH_k^\T.
\end{equation}
In this sense, we can view the first pass of the Kalman filter as an effective way to \emph{initialise} the EP parameters.
To account for the non-Gaussian likelihood in the update step we follow \citet{Nickish+Solin+Grigorievskiy:2018}, introducing an intermediary step in which the parameters of the \emph{sites},
$q^\textrm{site}_k(\vf_k) = \N(\vf_k \mid \vmu_k^{\mathrm{site}}, \MSigma_k^{\mathrm{site}})$,
are set via \emph{moment matching} and stored before continuing with the Kalman updates. 

This PEP formulation, with power $\alpha$, makes use of the fact that the required moments can be calculated via the derivatives of the log-normaliser, $\mathcal{L}_k$, of the tilted distribution \citep[see][]{seeger2005expectation}. Letting $\bm{\nabla}{\mathcal{L}}_k \in \R^{m}$ and $\bm{\nabla}^2{\mathcal{L}}_k \in \R^{m \times m}$ be the Jacobian and Hessian of $\mathcal{L}_k$ w.r.t. $\vmu_k^{\mathrm{cav}}$ respectively, this gives the following site update rule,
\begin{tcolorbox}
\paragraph{Power expectation propagation}
\begin{gather}
\begin{aligned} \label{eq:moment-match-filter}
&\mathcal{L}_k = \log \E_{\N(\vf_k \mid \vmu_k^{\mathrm{cav}},\MSigma_k^{\mathrm{cav}})}\big[p^\alpha(\vy_k \mid \vf_k) \big],  \\ 
&\MSigma_k^{\mathrm{site}} = -\alpha \left( \MSigma_k^{\mathrm{cav}} + \left(\bm{\nabla}^2{\mathcal{L}_k}\right)^{-1} \right) , \\ 
&\vmu_k^{\mathrm{site}} =  \vmu_k^{\mathrm{cav}} - \left(\bm{\nabla}^2{\mathcal{L}_k}\right)^{-1} \bm{\nabla}{\mathcal{L}_k} .
\end{aligned}
\end{gather}
\end{tcolorbox}
After the mean and covariance of our new likelihood approximation have been calculated, we proceed with a modified set of linear Kalman updates,
\begin{gather}
\begin{aligned} \label{eq:gaussian-update}
&\MS_k = \MH_k\MP_k^{\mathrm{pred}}\MH_k^\T + \MSigma_k^{\mathrm{site}}, \\
&\MK_k = \MP_k^{\mathrm{pred}} \MH_k^\T \MS_k^{-1}, \\
&\vm_k^{\mathrm{filt}} = \vm_k^{\mathrm{pred}} + \MK_k (\vmu_k^{\mathrm{site}} - \MH_k \vm_k^{\mathrm{pred}}), \\
& \MP_k^{\mathrm{filt}} = \MP_k^{\mathrm{pred}} - \MK_k \MS_k \MK_k^\T.
\end{aligned}
\end{gather}
As in \citet{wilkinson2019end}, we augment the RTS smoother with another moment matching step where the cavity distribution is calculated by removing (a fraction $\alpha$ of) the local site from the marginal smoothing distribution, \ie\ the posterior, $p(\vx_k \mid \vy_{1:n})=\N(\vx_k \mid \vm_k^{\textrm{post}},\MP_k^{\textrm{post}})$,
\begin{align} 
&\MSigma_k^{\mathrm{cav}} =  \big[ \hspace{-0.15em} {\big(\MH_k\MP_k^{\textrm{post}}\MH_k^\T\big)}^{-1} - \alpha {\left(\MSigma_k^{\mathrm{site}}\right)}^{-1} \big]^{-1}, \label{eq:moment-match-smoother} \\
&\vmu_k^{\mathrm{cav}} = \MSigma_k^{\mathrm{cav}} \big[ \hspace{-0.2em} {\big(\MH_k\MP_k^{\textrm{post}}\MH_k^\T\big)}^{-1} \MH_k\vm_k^{\textrm{post}} - \alpha {\left(\MSigma_k^{\mathrm{site}}\right)}^{-1} \vmu_k^{\mathrm{site}} \hspace{-0.1em} \big]. \nonumber
\end{align}
Moment matching is again performed via \cref{eq:moment-match-filter} using this new cavity. The site parameters, $\vmu_k^{\mathrm{site}}$, $\MSigma_k^{\mathrm{site}}$, are stored to be used on the next forward (filtering) pass. \cref{app:cavity} discusses methods for avoiding numerical issues that could occur due to the subtraction of covariance matrices in \cref{eq:moment-match-smoother}. \cref{alg:brief_algo} summarises the full learning algorithm, and \cref{app:marg-lik} describes how the marginal likelihood, $p(\vy\mid\vtheta)$, is computed to enable hyperparameter learning.

\subsection{Unifying Power\! EP and Extended Kalman\! Filtering} \label{sec:lin}

In the above inference scheme, a computational saving can be gained by noticing that when $\vh(\cdot, \cdot)$ is linear, $\mathcal{L}_k$ can be calculated in closed form. This fact has been exploited previously to aid inference in GP dynamical systems \cite{deisenroth2012expectation}. \cref{fig:Linearisation} demonstrates that such an approximation can be accurate when $\vh(\cdot,\cdot)$ is locally linear, or when the cavity variance is small. Using a first-order Taylor series expansion about the mean $\vmu_k^{\mathrm{cav}}$, we obtain
\begin{equation} \label{eq:taylor}
\vh(\vf_k,\vsigma_k) \approx \MJ_{\vf_k} (\vf_k-\vmu_k^{\mathrm{cav}}) + \vh(\vmu_k^{\mathrm{cav}}, \bm{0}) + \MJ_{\vsigma_k} \vsigma_k,
\end{equation}
a linear function of $\vf_k$ and $\vsigma_k \sim \N(\bm{0}, \MSigma_k)$, such that
$p(\vy_k \mid \vf_k) \approx \N(\vy_k \mid \hat{\vh}(\vf_k), \MJ_{\vsigma_k} \MSigma_k \MJ^\T_{\vsigma_k})$,
where $\hat{\vh}(\vf_k)=\MJ_{\vf_k} (\vf_k-\vmu_k^{\mathrm{cav}}) + \vh(\vmu_k^{\mathrm{cav}}, \bm{0})$. Here $\MJ_{\vf_k} = \MJ_{\vf}|_{\vmu_k^{\mathrm{cav}}, \bm{0}} \in \R^{d \times m}$ and $\MJ_{\vsigma_k} = \MJ_{\vsigma}|_{\vmu_k^{\mathrm{cav}}, \bm{0}} \in \R^{d \times d}$ are the Jacobians of $\vh(\cdot, \cdot)$ w.r.t.\ $\vf_k$ and $\vsigma_k$ evaluated at the mean, respectively. 

In order to frame approximate inference in the same setting as EP, we seek the site update rule implied by this linearisation. If $\MJ_{\vf}$ is invertible, then writing down such a rule would be trivial, but since this is not generally the case we instead use the EP moment matching steps, \cref{eq:moment-match-filter}, which give,  
\begin{align} 
\mathcal{L}_k &= \log \E_{\N(\vf_k \mid \vmu_k^{\mathrm{cav}},\MSigma_k^{\mathrm{cav}})}\big[\N^\alpha\big(\vy_k \mid \hat{\vh}(\vf_k), \MJ_{\vsigma_k} \MSigma_k \MJ^\T_{\vsigma_k} \big) \,\big] \nonumber \\
&= c + \log \N\big(\vy_k \mid \vh(\vmu_k^{\mathrm{cav}},\bm{0}) , \alpha^{-1} \hat{\MSigma}_k \big),
\label{eq:moment-match-linear}
\end{align}
where $\hat{\MSigma}_k= \MJ_{\vsigma_k} \MSigma_k \MJ^\T_{\vsigma_k} + \alpha \MJ_{\vf_k} \MSigma_k^{\mathrm{cav}} \MJ^\T_{\vf_k}$. Taking the derivatives of this log-Gaussian w.r.t.\ the cavity mean, we get
\begin{gather}
\begin{aligned} \label{eq:moment-match-derv}
\bm{\nabla}{\mathcal{L}}_k &= \diff{\mathcal{L}_k}{\vmu_k^{\mathrm{cav}}} = \alpha \MJ_{\vf_k}^\T \hat{\MSigma}_k^{-1}\vv_k , \\
\bm{\nabla}^2{\mathcal{L}_k} &= \diffIIvec{\mathcal{L}_k}{\vmu_k^{\mathrm{cav}}}{(\vmu_k^{\mathrm{cav}})^\T} = - \alpha \MJ^\T_{\vf_k} \hat{\MSigma}_k^{-1} \MJ_{\vf_k},
\end{aligned}
\end{gather}
where $\vv_k=\vy_k-\vh(\vmu_k^{\mathrm{cav}}, \bm{0})$. It is important to note that we have assumed the derivative of $\hat{\MSigma}_k$ to be zero, even though it depends on $\vmu_k^{\textrm{cav}}$. This assumption is crucial in ensuring that the updates are consistent, since it reflects the knowledge that the model is now linear (see \citet{deisenroth2012expectation} for detailed discussion). Now we update the site in closed form (\cref{app:site-updates} gives the derivation),
\begin{tcolorbox}
\paragraph{Extended expectation propagation}
\begin{gather}
\begin{aligned}
	\begin{split} \label{eq:site-update}
	\hspace{-1em}&\MSigma_k^{\mathrm{site}} = \left( \MJ^\T_{\vf_k} \left(\MJ_{\vsigma_k} \MSigma_k \MJ^\T_{\vsigma_k}\right)^{-1} \MJ_{\vf_k} \right)^{-1} , \\
	\hspace{-1em}&\vmu_k^{\mathrm{site}} = \vmu_k^{\mathrm{cav}} + \left( \MSigma_k^{\mathrm{site}} + \alpha \MSigma_k^{\mathrm{cav}} \right) \MJ^\T_{\vf_k}  \hat{\MSigma}_k^{-1} \vv_k. \hspace{-0.5em}
	\end{split}
\end{aligned}
\end{gather}
\end{tcolorbox}
The result when we use \cref{eq:site-update} (with $\alpha=1$) to modify the filter updates, \cref{eq:gaussian-update}, is \emph{exactly} the EKF (see \cref{app:derive-ekf} for the proof). Additionally, since these updates are now available in closed form, taking the limit $\alpha\rightarrow 0$ is now possible and avoids the matrix subtractions and inversions in \cref{eq:moment-match-smoother}, which can be costly and unstable. This is not possible prior to linearisation because the intractable integrals also depend on $\alpha$. \cref{app:algo} describes our full algorithm.
\renewcommand{\algorithmicensure}{\textbf{Return:}}
\begin{algorithm}[t]
	\caption{Sequential inference \& learning algorithm}
	\label{alg:brief_algo}
	\begin{algorithmic}
		\STATE {\bfseries Input:} $\{t_k,\vy_{k} \}^{n}_{k=1}$, $\vtheta_0$, $\alpha$, and learning rate $\rho$,
		\STATE \texttt{update\_rule} $\gets$ \cref{eq:moment-match-filter}, \cref{eq:site-update}, \cref{eq:param-update-sl} or \cref{eq:vi-update}
		\FOR{$i=1$ {\bfseries to} \texttt{num\_iters}}
		\STATE \mbox{build model, \cref{eq:gp-model}, with $\vtheta_{i-1}$. $\vm_{0}^{\mathrm{filt}},\MP_{0}^{\mathrm{filt}} \gets \bm{0}, \MP_\infty$}
		\FOR{ $k=1$ {\bfseries to} $n$}
		\STATE  $\vm_k^\textrm{pred},\MP_{k}^\textrm{pred}  \gets$ \texttt{predict} ($\vm_{k-1}^{\mathrm{filt}},\MP_{k-1}^{\mathrm{filt}}$)
		\IF{$i=1$} 
		\STATE  initialise $\vmu^{\mathrm{site}}_k$, $\MSigma^{\mathrm{site}}_k$ via \texttt{update\_rule} using $\alpha=1$ and $\vm_k^\textrm{pred},\MP_{k}^\textrm{pred}$ as cavity / posterior
		\ENDIF
		\STATE  \mbox{$\vm_k^{\mathrm{filt}},\MP_k^{\mathrm{filt}}  \gets \texttt{update}(\vm_k^\textrm{pred},\MP_{k}^\textrm{pred}, \vmu^{\mathrm{site}}_k, \MSigma^{\mathrm{site}}_k)$}
		\STATE $\ve_k = -\log p(\vy_k \mid \vy_{1:k-1}, \vtheta_{i-1})$ \comm{see \cref{app:marg-lik}} \\
		\ENDFOR
		\FOR{$k=n-1$ {\bfseries to} $1$}
		\STATE \mbox{$\vm_k^{\mathrm{post}},\MP_k^{\mathrm{post}} \gets \texttt{smooth}(\vm_{k+1}^{\mathrm{post}},\MP_{k+1}^{\mathrm{post}},\vm_k^{\mathrm{filt}}, \MP_k^{\mathrm{filt}})$}
		\STATE update $\vmu^{\mathrm{site}}_k$, $\MSigma^{\mathrm{site}}_k$ via \texttt{update\_rule}
		\ENDFOR
		\STATE $\vtheta_i = \vtheta_{i-1} + \rho \nabla_{\vtheta} \sum_k \ve_k$ \comm{update hyper.} \\
		\ENDFOR	
		\ENSURE posterior mean and covariance: $\vm^{\mathrm{post}}$, $\MP^{\mathrm{post}}$
	\end{algorithmic}
\end{algorithm}

\subsection{Power EP and the Unscented/GH Kalman Filters} \label{sec:gauss-filt}
We now consider the relationship between EP and general Gaussian filters, which use the likelihood approximation
\begin{align} \label{eq:gauss-filt}
p(\vy_k \mid \vf_k) \approx \N(\vy_k \mid &\vmu_k+\MC_k^\T (\MSigma^\textrm{cav}_k)^{-1}(\vf_k-\vmu^\textrm{cav}_k), \nonumber \\
 &\, \MS_k-\MC_k^\T (\MSigma^\textrm{cav}_k)^{-1} \MC_k),
\end{align}
where $\vmu_k$, $\MS_k$ and $\MC_k$ are the Kalman mean, innovation and cross-covariance terms respectively, given in \cref{app:kf_update}. \cref{eq:gauss-filt} amounts to \emph{statistical linear regression} \citep{Sarkka:2013} of $\vh(\vf_k,\vsigma_k)$. Letting $\vmu^\textrm{cav}_k=\MH_k\vm^\textrm{pred}_k$, $\MSigma^\textrm{cav}_k=\MH_k\MP^\textrm{pred}_k\MH_k^\T$ and using the Unscented transform / Gauss--Hermite to approximate $\vmu_k$, $\MS_k$ and $\MC_k$ results in the UKF / GHKF. This approximation has a similar form to the EKF (which uses \emph{analytical} linearisation, see \cref{fig:Linearisation} for comparison), and as in \cref{sec:lin} we can insert the Gaussian likelihood approximation into \cref{eq:moment-match-filter} to derive an iterated algorithm that matches the Gaussian filters on the first forward pass, but then refines the linearisation using EP style updates. This provides the following site update rule (see \cref{app:sl-site-updates}):
\begin{tcolorbox}
	\paragraph{\mbox{Statistically linearised expectation propagation}}
	\begin{gather}
	\begin{aligned} \label{eq:param-update-sl}
	\hspace{-1em}&\MSigma_k^{\mathrm{site}} =  - \alpha\MSigma_k^{\mathrm{cav}} + \left(\MOmega_k^\T \tilde{\MSigma}_k^{-1} \MOmega_k \right)^{-1} , \\
	\hspace{-1em}&\vmu_k^{\mathrm{site}} = \vmu_k^{\mathrm{cav}} + \left(\MOmega_k^\T \tilde{\MSigma}_k^{-1} \MOmega_k\right)^{-1} \MOmega_k^\T \tilde{\MSigma}_k^{-1} \vv_k .
	\end{aligned}
	\end{gather}
\end{tcolorbox}
where $\vv_k=\vy_k-\vmu_k$, $\tilde{\MSigma}_k= \MS_k + (\alpha-1) \MC_k^\T (\MSigma_k^{\mathrm{cav}})^{-1} \MC_k$,
\begin{multline} \label{eq:Omega}
\MOmega_k =\diff{\vmu_k}{\vmu_k^{\mathrm{cav}}} = \ \iint \vh(\vf_k, \vsigma_k) (\MSigma_k^{\mathrm{cav}})^{-1}(\vf_k - \vmu_k^{\textrm{cav}})  \\
\times \N(\vf_k \mid \vmu_k^{\mathrm{cav}}, \MSigma_k^{\mathrm{cav}}) \N(\vsigma_k \mid \bm{0}, \MSigma_k) \, \textrm{d}\vf_k \,\textrm{d}\vsigma_k.  
\end{multline}

\subsection{Nonlinear Kalman Smoothers}

Iterated versions of nonlinear filter-smoothers have been developed to address the fact that the forward prediction, $\N(\vx_k^\textrm{pred} \mid \vm_k^\textrm{pred}, \MP_k^\textrm{pred})$, may not be the optimal distribution about which to perform linearisation. It is argued that the posterior, $\N(\vx_k^\textrm{post} \mid \vm_k^{\textrm{post}},\MP_k^{\textrm{post}})$, obtained via smoothing, provides a better estimate of the region in which the likelihood affects the posterior \citep{garcia2015posterior}.

These iterated smoothers \citep{bell1994iterated} can be seen as special cases of the algorithms described in \cref{sec:lin} and \cref{sec:gauss-filt}, where the posterior is used to perform the linearisation in place of the cavity, \ie\ $\alpha=0$. The classical smoothers seek a linear approximation to the likelihood $p(\vy_k \mid \vf_k) \approx \N(\vy_k \mid \MB_k \vf_k + \vb_k, \ME_k)$ via a Taylor expansion, \cref{eq:taylor}, or SLR, \cref{eq:gauss-filt}, and then store parameters $\MB_k$, $\vb_k$, $\ME_k$ to be used during the next forward pass. 
Instead, we use the current posterior approximation to compute the site parameters via \cref{eq:site-update} or \cref{eq:param-update-sl}, which differs slightly from the standard presentation of these algorithms.
We argue that framing the Kalman smoothers as site update rules is beneficial in that it allows for direct comparison with EP, but also that introduction of the cavity is beneficial. The cavity may be a better distribution about which to linearise than the posterior, since it does not already include the effect of the local data. However, \cref{tbl:results} shows that setting $\alpha=0$ typically provides the best performance.

\subsection{Variational Inference with Natural Gradients}

Variational inference (VI) is an alternative to EP, often favoured due to its convergence guarantees and ease of implementation. If VI is formulated such that the variational parameters of the approximate posterior $q(\vf)$ are the likelihood (\ie site) mean and covariance, as in \cref{eq:fact-post}, then it can also be framed as a site update rule during Kalman smoothing \citep{chang2020}. This parametrisation is in fact the optimal one, as discussed in \citet{Opper+Archambeau:2009}, but is often avoided because the resulting optimisation problem is non-convex (instead it is common to declare a variational distribution over the full posterior, $q(\vf)=\N(\vm, \MK)$, and optimise $\vm$, $\MK$ with respect to the evidence lower bound. \citet{adam2020doubly} show how to perform \emph{natural gradient} VI under this parametrisation).

We present here the VI site update rule, based on conjugate-computation variational inference \citep[CVI,][]{khan2017}, in order to show their similarity to EP, and to enable direct comparison between the algorithms. CVI sidesteps the issues with the optimal parametrisation by showing that natural gradient VI can be performed via local site parameter updates that avoid directly differentiating the evidence lower bound. The updates can be written,
\begin{tcolorbox}
	\paragraph{Variational inference (with natural gradients)}
\begin{gather}
\begin{aligned}  \label{eq:vi-update}
&\tilde{\mathcal{L}}_k = \E_{\N(\vf_k \mid \vmu_k^{\mathrm{post}},\MSigma_k^{\mathrm{post}})}\big[\log p(\vy_k \mid \vf_k) \big], \\ 
&\MSigma_k^{\mathrm{site}} = - \left(\bm{\nabla}^2{\tilde{\mathcal{L}}_k}\right)^{-1}  , \\ 
&\vmu_k^{\mathrm{site}} =  \vmu_k^{\mathrm{post}} - \left(\bm{\nabla}^2{\tilde{\mathcal{L}}_k}\right)^{-1} \bm{\nabla}{\tilde{\mathcal{L}}_k} ,
\end{aligned}
\end{gather}
\end{tcolorbox}
where $\bm{\nabla}{\tilde{\mathcal{L}}}_k \in \R^{m}$ and $\bm{\nabla}^2{\tilde{\mathcal{L}}}_k \in \R^{m \times m}$ are the Jacobian and Hessian of $\tilde{\mathcal{L}}_k$ w.r.t. $\vmu_k^{\mathrm{post}}$ respectively.

\subsection{Spatio-Temporal Filtering and Smoothing}
\label{sec:s-t}
The methodology presented in the previous sections for temporal models directly lends itself to generalisations in spatio-temporal modelling. We consider a GP prior which is separable in the sequential (temporal) input $t$ and the remaining (spatial) input(s) $\vr$: $\kappa(\vr, t; \vr', t') = \kappa_\mathrm{r}(\vr,\vr')\,\kappa_\mathrm{t}(t,t')$.

Following \citet{Sarkka+Solin+Hartikainen:2013}, we extend the state $\vx(t)$ of the system via $m$ coupled temporal processes. These processes are associated with inducing points $\{\vr_{\mathrm{u},j}\}_{j=1}^m$ in the spatial domain. The measurement model matrix now projects the latent state at time $t_k$ from the inducing processes in the state to function space by, 
\begin{equation}
  \MH_k = [\MK_{\mathrm{f}_k,\mathrm{u}}\,\MK_{\mathrm{u},\mathrm{u}}^{-1}] \kron \MH_\mathrm{t}, 
\end{equation}
with Gram matrices $\MK_{\mathrm{f}_k,\mathrm{u}} = \kappa_\mathrm{r}(\vr_k,\vr_{\mathrm{u},j})$ and $\MK_{\mathrm{u},\mathrm{u}} = \kappa_\mathrm{r}(\vr_{\mathrm{u},j},\vr_{\mathrm{u},j})$ for $j=1,2,\ldots,m$, where $\MH_{\mathrm{t}}$ is the measurement model matrix for the GP prior. If the data lies on a fixed set of spatial points $\{\vr_j\}_{j=1}^m$ (an irregular grid of $m$ points), the above expression simplifies to $\MH_k = \MI_m \kron \MH_t$ and the models becomes exact \citep[see,][for further details and discussion, also covering non-separable models]{Hartikainen:2013,Solin:2016}.

\subsection{Fast Implementation Using JAX}

Sequential inference in GPs is extremely efficient, however optimising the model hyperparameters involves differentiating functions with large loops. When using automatic differentiation this typically results in a massive computational graph with large compilation overheads, memory usage and runtime. Previous approaches have avoided this issue either by using finite differences \citep{Nickish+Solin+Grigorievskiy:2018}, which are slow when the number of parameters is large, or by reformulating the model to exploit linear algebra methods applicable to sparse precision matrices \citep{durrande2019}.

We utilise the following features of the differential programming Python framework, JAX \citep{jax2018}: \emph{(i)}~we avoid `unrolling' of for-loops, \ie\ instead of building a large graph of repeated operations, a smaller graph is recursively called, reducing the compilation overhead and memory, \emph{(ii)}~we just-in-time (JIT) compile the loops, to avoid the cost of graph retracing, \emph{(iii)}~we use accelerated linear algebra (XLA) to speed up the underlying filtering/smoothing operations. Combined, these features result in an extremely fast implementation, and based on this we provide a fully featured temporal GP framework with all models and inference methods, available at \url{https://github.com/AaltoML/kalman-jax}.

\begin{table*}[t]
	\caption{Normalised negative log predictive density (NLPD) results with 10-fold cross-validation. Smaller is better. Blank entries (---) represent scenarios where the method does not scale to the size of the task. EEP, UEP, and GHEP are the iterated smoothers with linearisation. \textsc{EP(u)} and \textsc{EP(gh)} are state space EP, where the intractable moment matching is performed via the Unscented transform or Gauss--Hermite, respectively. Linearisation performs poorly on the heteroscedastic noise task, however EEP performs well on the audio task since it is the only method capable of maintaining full site cross-covariance terms without compromising stability. The state space methods are able to match the performance of the non-sequential (batch) EP and VGP baselines.}
	\label{tbl:results}
	\tikzstyle{every picture}+=[remember picture]
	\tikzstyle{na} = [baseline=-.5ex]
	\begin{center}
		{\fontsize{7.15pt}{10.5pt} 
			\selectfont
			\setlength{\tabcolsep}{0pt}
			\newlength{\tblw}
			\setlength{\tblw}{0.12\textwidth}
			\begin{sc}
				\begin{tabularx}{\textwidth}{p{1.65em} l @{\extracolsep{\fill}} C{\tblw} C{\tblw} C{\tblw} C{\tblw} C{\tblw} C{\tblw} C{\tblw}}
					\toprule
					&& Motorcycle & Coal & Banana & Binary & Audio & Airline & Rainforest \\
					\midrule
					\multicolumn{2}{l}{\# data points} & 133 & 333 & 400 & 10k & 22k & 36k & 125k \\
					\multicolumn{2}{l}{Input dimension} & 1 & 1 & 2 & 1 & 1 & 1 & 2 \\
					\multicolumn{2}{l}{Likelihood} & Heteroscedastic & Poisson & Bernoulli & Bernoulli & Product & Poisson & Poisson \\
					\midrule 
					\rowcolor{gray!10}                  
					&EEP ($\alpha=1$) &0.855$\pm$0.25 &0.922$\pm$0.11 &0.228$\pm$0.07 &0.536$\pm$0.01 &$-$0.433$\pm$0.04 &0.142$\pm$0.01 &0.325$\pm$0.01 \\
					\rowcolor{gray!10}
					&EEP ($\alpha=0.5$) &0.855$\pm$0.25 &0.922$\pm$0.11 &0.228$\pm$0.07 &0.536$\pm$0.01 &$-$0.499$\pm$0.03 &0.142$\pm$0.01 &0.321$\pm$0.01 \\
					\rowcolor{gray!10} 
					&EEP ($\alpha=0$) / EKS &0.855$\pm$0.25 &0.922$\pm$0.11 &0.229$\pm$0.07 &0.537$\pm$0.01 &\textbf{$-$0.570$\pm$0.04} &0.142$\pm$0.01 &\textbf{0.309$\pm$0.01} \\
					&UEP ($\alpha=1$) &0.745$\pm$0.28 &0.922$\pm$0.11 &0.217$\pm$0.08  &0.536$\pm$0.01 &$-$0.471$\pm$0.02 &0.142$\pm$0.01 & --- \\
					\tikz[na]\node[coordinate] (n1) {};
					&UEP ($\alpha=0.5$) &0.745$\pm$0.28 &0.922$\pm$0.11 &0.217$\pm$0.08 &0.536$\pm$0.01   &$-$0.474$\pm$0.02 &0.142$\pm$0.01 &--- \\
					&UEP ($\alpha=0$) / UKS &0.745$\pm$0.28 &0.922$\pm$0.11 &0.217$\pm$0.08  &0.536$\pm$0.01 &$-$0.484$\pm$0.02 &0.142$\pm$0.01 &--- \\
					\rowcolor{gray!10} 
					&GHEP ($\alpha=1$) &0.750$\pm$0.26  &0.922$\pm$0.11&0.217$\pm$0.08 &0.536$\pm$0.01 & --- &0.142$\pm$0.01 &--- \\
					\rowcolor{gray!10}
					&GHEP ($\alpha=0.5$) &0.747$\pm$0.27 &0.922$\pm$0.11 &0.217$\pm$0.08 &0.536$\pm$0.01 &--- &0.142$\pm$0.01 &--- \\
					\rowcolor{gray!10} 
					&GHEP\! ($\alpha=0$)\! /\! GHKS &0.746$\pm$0.27 &0.922$\pm$0.11 &0.217$\pm$0.08 &0.536$\pm$0.01 & --- &0.142$\pm$0.01 &--- \\
					&EP(u) ($\alpha=1$) & 0.696$\pm$0.59 &0.922$\pm$0.11 &0.217$\pm$0.08 &0.536$\pm$0.01 &$-$0.321$\pm$0.15 &0.143$\pm$0.01 &      --- \\
					&EP(u) ($\alpha=0.5$) &0.479$\pm$0.30 &0.922$\pm$0.11 &0.217$\pm$0.08 &0.536$\pm$0.01&$-$0.327$\pm$0.20 &0.143$\pm$0.01 &      --- \\
					&EP(u) ($\alpha\approx 0$) &0.465$\pm$0.29 &0.924$\pm$0.11 &0.222$\pm$0.08 &0.536$\pm$0.01 &0.011$\pm$0.30 &0.143$\pm$0.01 &      --- \\
					\rowcolor{gray!10} \tikz[na]\node[coordinate] (n2) {};
					&EP(gh) ($\alpha=1$) & 0.569$\pm$0.41 &0.922$\pm$0.11 &0.217$\pm$0.08 &0.536$\pm$0.01 &      --- &0.142$\pm$0.01 &      --- \\
					\rowcolor{gray!10}
					&EP(gh) ($\alpha=0.5$) &0.531$\pm$0.38 &0.922$\pm$0.11 &0.217$\pm$0.08 &0.536$\pm$0.01 &      --- &0.142$\pm$0.01 &      --- \\
					\rowcolor{gray!10} 
					&EP(gh) ($\alpha\approx 0$) & \textbf{0.444$\pm$0.32} &0.922$\pm$0.11 &0.217$\pm$0.08 &0.536$\pm$0.01 &      --- &0.142$\pm$0.01 &      --- \\
					&VI(u)  & \textbf{0.444$\pm$0.31} &0.922$\pm$0.11 &0.217$\pm$0.08 &0.536$\pm$0.01 &$-$0.204$\pm$0.02 &0.142$\pm$0.01&--- \\
					\tikz[na]\node[coordinate] (n3) {};
					&VI(gh) &0.495$\pm$0.34 &0.922$\pm$0.11 &0.217$\pm$0.08  &0.536$\pm$0.01 & --- &0.142$\pm$0.01 &--- \\
					\hline
					\addlinespace[0.15em]
					\tikz[na]\node[coordinate] (n4) {};
					&EP(batch, gh) & 0.441$\pm$0.30 & 0.922$\pm$0.11 & 0.216$\pm$0.10 & ---& ---& ---& ---\\
					&VGP(batch, gh) & 0.444$\pm$0.30 & 0.922$\pm$0.11 & 0.219$\pm$0.09 & ---& ---& ---& ---\\
					\bottomrule
				\end{tabularx}
			\end{sc}						
		}	
	\end{center}
	\begin{tikzpicture}[overlay]
        \node[rotate=90,yshift=-.5em] at (n1) {\tiny \textsc{Linearisation}};
        \node[rotate=90,yshift=-.5em,xshift=.5em] at (n2) {\tiny \textsc{Moment Match}};
        \node[rotate=90,yshift=-.5em,xshift=.5em] at (n3) {\tiny VI};
        \node[rotate=90,yshift=-.5em,xshift=-.5em] at (n4) {\tiny \textsc{Basel.}};        
	\end{tikzpicture}
	\vspace*{-2em}
\end{table*}

\section{Empirical Analysis}
\label{sec:results}
\begin{figure}[t!]
	\centering\tiny
	\newcommand{\mycaption}[1]{\rotatebox{90}{\parbox[c][2em][c]{1.2\figureheight}{\centering\footnotesize #1}}}
	\pgfplotsset{yticklabel style={rotate=90}, ylabel style={yshift=0pt},scale only axis,axis on top,clip=false}
	\setlength{\figurewidth}{.8\columnwidth}
	\setlength{\figureheight}{.4\figurewidth}
	\begin{minipage}[t]{\columnwidth}
	  \raggedright
	  \pgfplotsset{ignore legend}
	  \mycaption{(a)~Motorcycle}\phantomsubcaption~~	  	  
	  \input{./fig/mcycle.tex}
	  \label{fig:mcycle}        
	\end{minipage}\\
	\begin{subfigure}[t]{\columnwidth}
      \raggedright
	  \mycaption{(b)~Coal mining}\phantomsubcaption~~
      \input{./fig/coal.tex}
	  \label{fig:coal}    
	\end{subfigure}\\
	\begin{subfigure}[t]{\columnwidth}
	  \raggedright
	  \pgfplotsset{ignore legend}
	  \mycaption{(c)~Airline accidents}\phantomsubcaption~~
%
%
\begin{tikzpicture}

\begin{axis}[%
axis on top,
xmin=1919.49404133235,
xmax=2018.05390387313,
xtick={1920, 1930, 1940, 1950, 1960, 1970, 1980, 1990, 2000, 2010, 2020},
xlabel style={font=\color{white!15!black}},
xlabel={Time (years)},
ymin=-0.0277777777777778,
ymax=38.0277777777778,
ylabel style={font=\color{white!15!black}},
ylabel={Accident intensity, $\lambda(t)$},
axis background/.style={fill=white},
legend style={legend cell align=left, align=left, draw=white!15!black},
width=\figurewidth,
height=\figureheight
]
\addplot [forget plot] graphics [xmin=1919.49404133235, xmax=2018.05390387313, ymin=-0.0277777777777778, ymax=38.0277777777778] {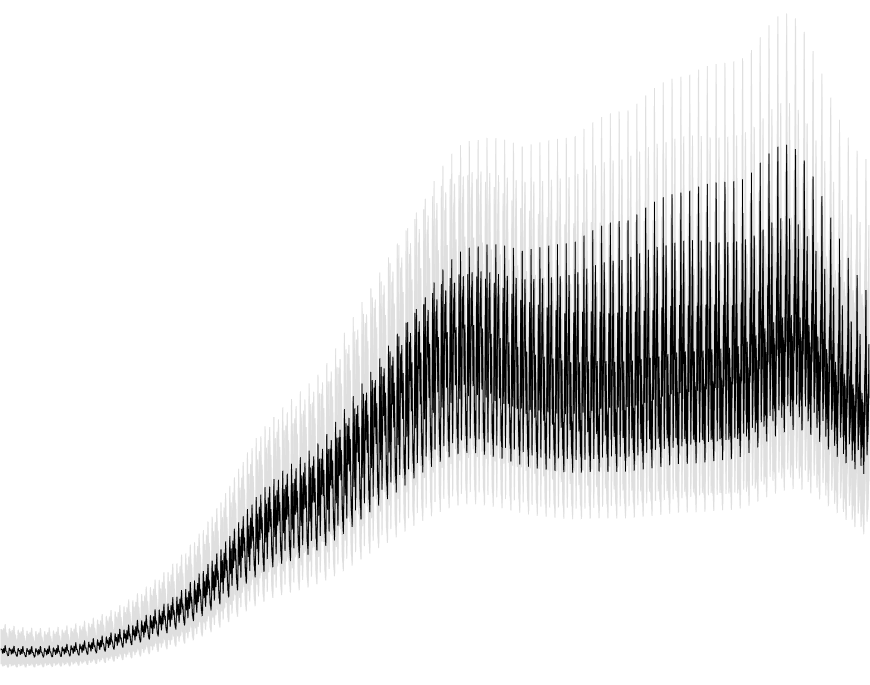};
\end{axis}
\end{tikzpicture}%
	  \label{fig:airline}  
	\end{subfigure}\\
	\begin{subfigure}[t]{\columnwidth}
	  \raggedright
	  \mycaption{(d)~Rainforest}\phantomsubcaption~~	  
%
%
\begin{tikzpicture}

\begin{axis}[%
axis on top,
xmin=-1.6410371763936,
xmax=1001.64103717639,
xlabel style={font=\color{white!15!black}},
xlabel={$t$: spatial dimension 1 (metres)},
ymin=-1.64433243176789,
ymax=501.644332431768,
ylabel style={font=\color{white!15!black}},
ylabel={$r$: spatial dimension 2 (metres)},
axis background/.style={fill=white},
legend style={legend cell align=left, align=left, draw=white!15!black,fill=white},
width=\figurewidth,
height=\figureheight,
legend image post style={scale=0}
]
\addlegendimage{only marks,no markers}
\addlegendentry{ log-intensity, $\log \lambda(r,t)$}
\addplot [forget plot] graphics [xmin=-1.6410371763936, xmax=1001.64103717639, ymin=-1.64433243176789, ymax=501.644332431768] {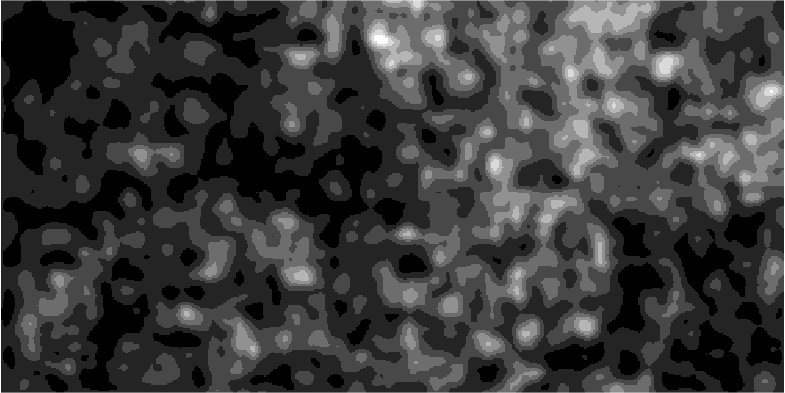};

\end{axis}
\end{tikzpicture}%
	  \label{fig:rainforest}  
	\end{subfigure}
	\vspace*{-2em}
	\caption{Examples of non-conjugate GP models. (a)~In the motorcycle task (heteroscedastic noise), EP is capable of modelling the time-varying noise component. The log-Gaussian Cox processes (b)--(d) are well approximated via linearisation, and iterating improves the match to the EP posterior.}
	\label{fig:results}
	\vspace*{-2em}
\end{figure}

\cref{tbl:results} examines the performance of 17 methods from our GP framework on 7 benchmark tasks of varying data size and model complexity. Blanks (---) in the table represent scenarios where the method does not scale practically to the size of the task. First we demonstrate that state space approximate inference schemes are competitive with two state-of-the-art baseline methods on three small datasets. We compare against batch EP (see \cref{sec:EP}) and a variational GP (VGP, \citealp{Opper+Archambeau:2009}, with order $n+n^2$ parameters to ensure convexity of the objective, as in GPflow, \citealp{GPflow:2017}). We then compare our methods on four large data tasks to which the baselines are not applicable. Note that sparse variants of EP and VGP are not suited to these long time series containing high-frequency behaviour (\eg, \cref{fig:airline}) that cannot be summarised by a few thousand inducing points (see \cref{sec:intro} for discussion).

We evaluate all methods via negative log predictive density (NLPD) using 10-fold cross-validation, with each method run for 250 iterations (baselines are run until convergence). Gauss--Hermite integration uses $20^q$ cubature points, whereas the Unscented transform uses $2q^2+1$ \citep[we use the symmetric 5\textsuperscript{th} order cubature rule, \ie UT5,][]{kokkala2016a, mcnamee1967construction}, where $q$ is the dimensionality of the integral (typically the number of GPs that are nonlinearly combined in the likelihood). For standard EP, where a power of zero is not possible, we set $\alpha=0.01$.

We optimise the model hyperparameters by maximising the marginal likelihood $p(\vy \mid \vtheta)$ separately for each method (see \cref{app:marg-lik} for details), hence the results in \cref{tbl:results} are affected by both training and inference, demonstrating their applicability as practical machine learning algorithms. The baseline methods scale as $\mathcal{O}(n^3)$, while all the sequential schemes scale as $\mathcal{O}(ns^3)$.

\paragraph{Results} \cref{tbl:results} confirms it is possible to achieve state-of-the-art performance with sequential inference. The log-Gaussian Cox process and classification experiments return consistent results across all methods. However, Audio and Rainforest involve multidimensional sites, making them difficult tasks. In such cases, EEP performs well because it is the only method capable of maintaining full site covariance terms whilst remaining stable. Statistical linearisation suffers less from a reduction of cubature points than EP moment matching or VI updates, as shown by the performance of UEP on Audio. EP generally performed well, but EEP matches its performance sometimes whilst being the only method applicable to the Rainforest task. In other cases, cubature methods outperform linearisation, particularly on the Motorcycle task.

\begin{table}[t]
	\caption{Run times in seconds (mean across 10 runs). We report time to evaluate the marginal likelihood on the forward pass \emph{and} perform the site updates on the smoothing pass.}
	\label{tbl:runtime}
	
	\newcommand{\rotated}[1]{\rotatebox[origin=c]{90}{\begin{minipage}{2.5cm}#1\end{minipage}}}
	
	\resizebox{\columnwidth}{!}{
				\begin{tabular}{lccccccr}
					\toprule
					& \rotated{{\sc Motorcycle} \\ (Heteroscedastic)}  
					& \rotated{{\sc Coal} \\ (Poisson)} 
					& \rotated{{\sc Banana} \\ (Bernoulli)} 
					& \rotated{{\sc Binary} \\ (Bernoulli)}
					& \rotated{{\sc Audio} \\ (Product)} 					
					& \rotated{{\sc Airline} \\ (Poisson)} 	
					& \rotated{{\sc Rainforest} \\ (Poisson)} \\
					\midrule           
					EEP      &0.015 &0.013 &0.135 &0.100 & 1.176 &23.941 &37.441 \\             
					UEP     &0.017 &0.015 &0.144 &0.113 & 1.661 & 24.583 &   --- \\               
					GHEP   &0.020 &0.016 &0.146 &0.120 &---     & 23.709 &   --- \\              
					EP(U)   &0.018 &0.015 &0.143 &0.108 & 1.713 &23.492 &   --- \\             
				 	EP(GH) &0.019 &0.016 &0.150 &0.127 &---     &23.777 &   --- \\    
				 	VI(U)    &0.017 &0.017 &0.142 &0.098 &1.619 &23.796 &   --- \\    
				 	VI(GH)  &0.018 &0.016 &0.145 &0.123 &---     &23.611 &   --- \\                
					\midrule
					Time steps & 133 & 333 & 400 & 10000 & 22050 & 35959 & 500 \\
					State dim. & 6 & 3 & 45 & 4 & 15 & 59 & 500\\
					\bottomrule
				\end{tabular}}
				\vspace*{-1em}
\end{table}

\paragraph{Motorcycle (heteroscedastic noise)}
The motorcycle crash dataset \citep{Silverman:1985} contains 131 non-uniformly spaced measurements from an accelerometer placed on a motorcycle helmet during impact, over a period of 60~ms. It is a challenging benchmark \citep[\eg,][]{Tolvanen:2014}, due to the heteroscedastic noise variance. We model both the process itself and the measurement noise scale with independent GP priors with Mat\'ern-$\nicefrac{3}{2}$ kernels: 
$y_k \mid f_k^{(1)},f_k^{(2)} \sim \N(y_{k} \mid f^{(1)}(t_{k}),[\phi(f^{(2)}(t_k))]^2),$
with softplus link function $\phi(f) = \log(1+e^f)$ to ensure positive noise scale. The full EP and VGP baselines were implemented and hand-tailored for this task. For VGP, we used GPflow~2 \citep{GPflow:2017} with a custom model. VI and EP ($\alpha \approx 0$) performed well, however the linearisation-based methods failed to capture the time-varying noise (see \cref{sec:app-models} for discussion).

\paragraph{Coal (log-Gaussian Cox process)}
The coal mining disaster dataset \citep{Vanhatalo+Riihimaki+Hartikainen+Jylanki+Tolvanen+Vehtari:2013} contains 191 explosions that killed ten or more men in Britain between 1851--1962. We use a log-Gaussian Cox process, \ie\ an inhomogeneous Poisson process (approximated with a Poisson likelihood for $n=333$ equal time interval bins). We use a Mat\'ern-$\nicefrac{5}{2}$ GP prior with likelihood 
$ p(\vy \mid \vf) \approx \prod_{k=1}^n \mathrm{Poisson}(y_k \mid \exp(f(\hat{t}_k)))$,
where $\hat{t}_k$ is the bin coordinate and $y_k$ the number of disasters in the bin. This model reaches posterior consistency in the limit of bin width going to zero \citep{Tokdar+Ghosh:2007}. Since linearisation requires a continuous likelihood, we approximate the discrete Poisson with a Gaussian by noticing that its first two moments are equal to the intensity $\lambda(t) = \exp(f(t))$, giving $y_k \mid f_k \stackrel{\text{approx.}}{\sim} \N(y_k \mid \lambda(\hat{t}_k), \lambda(\hat{t}_k))$. See \cref{sec:app-models} for details.

\begin{figure*}[t]
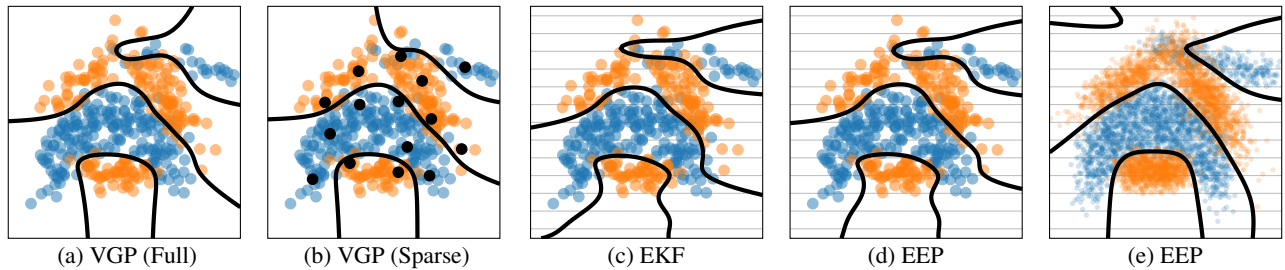

  \centering
  \pgfplotsset{yticklabel style={rotate=90}, ylabel style={yshift=0pt},scale only axis,axis on top,clip=true,clip marker paths=true}
  \setlength{\figurewidth}{.18\textwidth}
  \setlength{\figureheight}{\figurewidth}
  \begin{subfigure}[b]{.19\textwidth}
    \centering
    \input{./fig/banana-vgp.tex}
    \caption{VGP (Full)}
  \end{subfigure}
  \hfill
  \begin{subfigure}[b]{.19\textwidth}
    \centering
    \input{./fig/banana-svgp.tex}
    \caption{VGP (Sparse)}
  \end{subfigure}
  \hfill
  \begin{subfigure}[b]{.19\textwidth}
    \centering
    \input{./fig/banana-ekf.tex}
    \caption{EKF}
  \end{subfigure}
  \hfill
  \begin{subfigure}[b]{.19\textwidth}
    \centering
    \input{./fig/banana-ek-pep.tex}
    \caption{EEP}    
  \end{subfigure}
  \hfill
  \begin{subfigure}[b]{.19\textwidth}
    \centering
    \input{./fig/banana-ekf-large.tex}
    \caption{EEP}  
    \label{fig:ek-ep-large}  
  \end{subfigure}
  \vspace*{-1em}
  \caption{Inference schemes on the two-dimensional {\em Banana} classification task. The coloured points represent training data and the black lines are decision boundaries. (a)~is the baseline variational GP method (VGP in \cref{tbl:results}). (b)~shows the sparse variant of the VGP baseline (Generalized FITC) with $15$ inducing points (black dots). In (c)--(e), the vertical dimension is treated as the `spatial' input ($m=15$ inducing points shown with lines) and the horizontal as the sequential (`temporal') dimension. Our formulation using the EKF (c) works well, but is further improved by the EP-like iteration in (d). In (e), the method is applied to the full data set ($n=5400$).}
  \label{fig:banana}
  \vspace*{-1em}
\end{figure*}

\paragraph{Airline (log-Gaussian Cox process)} 
The airline accidents data \citep{Nickish+Solin+Grigorievskiy:2018} consists of 1210 dates of commercial airline accidents between 1919--2017. We use a log-Gaussian Cox process with bin width of one day, leading to $n = 35{,}959$ observations. The prior has multiple components, $\kappa(t,t') = \kappa(t,t')^{\nu=\nicefrac{5}{2}}_{\text{Mat.}}  + \kappa(t,t')_{\text{per.}}^{1\,\text{year}} \kappa(t,t')^{\nu=\nicefrac{3}{2}}_{\text{Mat.}} + \kappa(t,t')_{\text{per.}}^{1\,\text{week}} \kappa(t,t')^{\nu=\nicefrac{3}{2}}_{\text{Mat.}} $, capturing a long-term trend, time-of-year variation (with decay), and day-of-week variation (with decay). The state dimension is $s=59$. 

\paragraph{Binary (1D classification)}
As a 1D classification task, we create a long binary time series, $n = 10{,}000$, using the generating function $y(t) = \text{sign} \{ \frac{12 \sin(4 \pi t)} {0.25 \pi t +1} + \sigma_t\}$, with $\sigma_t \sim \N(0, 0.25^2)$. Our GP prior has a Mat\'ern-$\nicefrac{7}{2}$ kernel, $s=4$, and the sigmoid function $\psi(f) = (1 +e^{-f})^{-1}$ maps $\R \mapsto [0,1]$ (logit classification). See \cref{sec:app-models} for derivation of approximate continuous state observation model, $h(f_k,\sigma_k)= \psi(f_k) + \sqrt{\psi(f_k)(1-\psi(f_k))}\sigma_k$.

\paragraph{Audio (product of GPs)} 
We apply a simplified version of the Gaussian Time-Frequency model from \citet{wilkinson2019end} to half a second of human speech, sampled at 44.1~kHz, $n=22{,}050$. The prior consists of 3 quasi-periodic ($\kappa_\textrm{exp}(t,t') \kappa_\textrm{cos}(t,t')$) `subband' GPs, and 3 smooth ($\kappa_\textrm{Mat-\nicefrac{5}{2}}(t,t')$) `amplitude' GPs. The likelihood consists of a sum of the product of these processes with additive noise and a softplus mapping $\phi(\cdot)$ for the positive amplitudes: $y_k \mid \vf_k \sim \N(\sum_{i=1}^3 f_{i,k}^\textrm{sub.} \phi(f_{i,k}^\textrm{amp.}), \sigma^2_k)$. The nonlinear interaction of 6 GPs ($s=15$) in the likelihood makes this a challenging task. EEP performs best since it is capable of maintaining full site covariance terms without compromising stability. UEP outperforms EP and VI since statistical linearisation is still accurate when using few cubature points.

\subsection{Spatio-Temporal Models}
As presented in \cref{sec:s-t}, the sequential inference schemes are also applicable to spatio-temporal problems. We illustrate this via two spatial problems, treating one spatial input as the sequential dimension (`time') and the other as `space'.

\paragraph{Banana (2D classification)}
The banana data set, $n=400$, is a common 2D classification benchmark \citep[\eg,][]{Hensman+Matthews+Ghahramani:2015}. We use the logit likelihood with a separable space-time kernel: $\kappa(r,t;r',t') =  \kappa(t,t')_{\text{Mat.}}^{\nu=\nicefrac{5}{2}} \kappa(r,r')^{\nu=\nicefrac{5}{2}}_{\text{Mat.}}$. The vertical dimension is treated as space $r$ and the horizontal as the sequential (`temporal') dimension $t$. We use $m=15$ inducing points in $r$ (see \cref{sec:s-t}), visualised by lines in \cref{fig:banana}(c)--(e). The state dimension is $s = 3m = 45$.
\cref{fig:banana} shows that the EKF provides a similar solution to the VGP baseline of \citet{Hensman+Matthews+Ghahramani:2015}, and an even closer match is obtained by EEP (3 iterations). The forward and backward passes are visualised in \cref{fig:banana-ek}. The method is also applicable to the larger ($n=5400$) version of the data set (\cref{fig:ek-ep-large}).

\paragraph{Rainforest (2D log-Gaussian Cox process)}
We study the density of a single tree species, \emph{Trichilia tuberculata}, from a 1000$\,$m $\times$ 500$\,$m region of a rainforest in Panama \citep{rainforest,Condit:2005,Hubbell:1999}. We segment the space into $4\,\text{m}^2$ bins, giving a $500\times 250$ grid with $125{,}000$ data points ($n=500$ time steps), and use a log-Gaussian Cox process (\cref{fig:rainforest}). The space-time GP has a separable Mat\'ern-$\nicefrac{3}{2}$ kernel. We \emph{do not} use a sparse approximation in $r$, instead we have $m=250$ temporal processes, so $s=2m=500$.

\paragraph{Run Times} 
\cref{tbl:runtime} compares time taken for all methods to make a single training step on a MacBook Pro with 2.3~GHz Intel Core i5 and 16~GB RAM using JAX. For tasks with one-dimensional sites all methods are similar, however EEP is faster than the cubature methods for the Audio task which involves 6-dimensional sites. The gridded data of the Rainforest task requires a 250-dimensional site parameter update which is impractical for most methods. Conversely, in the Banana task data points are handled one by one, such that only one-dimensional updates are required.

\section{Discussion and Conclusions}
\label{sec:discussion}
We argue that development of methods capable of naturally handling sequential data is crucial to extend the applicability of GPs beyond short time series. EP was originally inspired by, and derived from, Kalman filtering and here we make the case that a return to sequential methods is desirable for large spatio-temporal problems. We present a flexible and efficient framework for sequential learning that encompasses many state-of-the-art approximate inference schemes, whilst also illuminating the connections between modern day inference methods and traditional filtering approaches.

Our theoretical contributions confirm that using linearisation in place of EP moment matching results in iterated algorithms that exactly match the classical nonlinear Kalman filters on the first pass, and also generalise the classical smoothers by refining the linearisations via multiple passes through the data. These algorithms are fast and scale to high-dimensional spatio-temporal problems more effectively than EP and VI. The methods based on Taylor series approximations only require one evaluation of the likelihood (and its Jacobian) for each data point, as opposed to cubature methods, and these algorithms make it particularly straightforward to prototype and implement new likelihood models.

We provide a detailed examination of the different properties of all these methods on five time series and two spatial tasks, showing that the state space framework for GPs is applicable beyond one-dimensional problems. We have also highlighted the scenarios in which such methods might fail: linearisation is a poor approximation when the cavities are diffuse (high variance) and the likelihood is highly nonlinear, but cubature methods do not scale well to high dimensions.

\section*{Acknowledgements}
We acknowledge funding from the Academy of Finland (grant numbers 308640 and 324345) and from Innovation Fund Denmark (grant number 8057-00036A). Our results were obtained using computational resources provided by the Aalto Science-IT project. Thanks to S.\ T.\ John for help implementing the VGP baseline for the heteroscedastic noise model.

\bibliography{bibliography}
\bibliographystyle{icml2020}

%
%
%

\clearpage

\appendix

\newcommand{\nipstitle}[1]{{%
    \phantomsection\hsize\textwidth\linewidth\hsize%
    \vskip 0.1in%
    \toptitlebar%
    \begin{minipage}{\textwidth}%
        \centering{\Large\bf #1\par}%
    \end{minipage}%
    \bottomtitlebar%
    \addcontentsline{toc}{section}{#1}%
}}

\clearpage
\normalsize\twocolumn[


\nipstitle{
    {Supplementary Material:} \\
    State Space Expectation Propagation
}]
\pagestyle{empty}

\section{Nomenclature}\label{app:nomenclature}
Vectors: bold lowercase. Matrices: bold uppercase. \\
\setlength{\tabcolsep}{1pt}
\begin{tabularx}{\columnwidth}{ll}
\toprule
Symbol & Description \\
\midrule
$n$ & Number of time steps \\
$m$ & Number of latent functions / processes \\
$s$ & State dimensionality \\
$d$ & Output dimensionality \\
$t \in \R$ & Time (input) \\
$\vr$ & Space (input, of arbitrary dimension) \\
$k$ & Time index, $t_k$, $k=1,\ldots,n$ \\
$\vy_k \in \R^d$ & Observation (output) \\
$\vy \in \R^{d\times n}$ & Collection of outputs, $(\vy_1, \vy_2, \ldots, \vy_n)$ \\
$\vtheta$ & Vector of model (hyper)parameters \\
$\kappa(t,t')$ & Covariance function (kernel) \\
$\MK(t,t')$ & Multi-output covariance function \\
$\mu(t)$ & Mean function \\
$\vmu(t)$ & Multi-output mean function \\
$\vsigma_k$ & Measurement noise \\
$\MSigma_k$ & Measurement noise covariance \\
$f(t): \R \to \R$ & Latent function (Gaussian process) \\
$\vf(t): \R \to \R^m$ & Vector of latent functions, $\vf_k=\vf(t_k)$ \\
$\vf \in \R^{m\times n}$ & Collection of latents, $(\vf(t_1),\ldots, \vf(t_n))$ \\
$\MH_k \in \R^{m \times s}$ & State$\rightarrow$function mapping \\
$\vh(\vf_k, \vsigma_k)$ & Measurement model $(\R^{m},\R^{d}) \to \R^{d}$ \\
$\vx(t): \R \to \R^s$ & State vector, $\vf(t) = \MH_k\vx(t)$ \\
$\vx_k \in \R^s$ & State variable, $\vx_k = \vx(t_k) \sim \N(\vm_k, \MP_k)$ \\
$\MF \in \R^{s\times s}$ & Feedback matrix (continuous) \\
$\ML \in \R^{s \times v}$ & Noise effect matrix (continuous) \\
$\MQ_\mathrm{c} \in \R^{v \times v}$ & White noise spectral density (continuous) \\
$\MA_k \in \R^{s\times s}$ & Dynamic model (discrete) \\
$\vq_k \in \R^s $ & State space process noise (discrete) \\
$\MQ_k \in \R^{s\times s}$ & Process noise covariance (discrete) \\
$\MP_\infty \in \R^{s\times s}$ & Stationary state covariance (prior) \\
$\vm_k \in \R^{s\times 1}$ & State mean \\
$\MP_k \in \R^{s\times s}$ & State covariance \\
$\MK_k \in \R^{s\times d}$ & Kalman gain \\
$\MG_k \in \R^{s\times s}$ & Smoother gain \\
$\MJ_{\vf} \in \R^{d \times m}$ & Jacobian of $\vh$ w.r.t $\vf_k$ \\
$\MJ_{\vsigma} \in \R^{d \times d}$ & Jacobian of $\vh$ w.r.t $\vsigma_k$ \\
$\alpha$ & EP power / fraction \\
$\mathcal{L}_k$ & log-normaliser of true posterior update \\
$q^\textrm{site}_k(\vf_k)$ & EP site (approximate likelihood) \\
 & $q^\textrm{site}_k(\vf_k)\sim\N(\vmu_k^\textrm{site},\MSigma_k^\textrm{site})$ \\
$q^\textrm{cav}_k(\vf_k)$ & EP cavity (leave-one-out posterior) \\
 & $q^\textrm{cav}_k(\vf_k)\sim\N(\vmu_k^\textrm{cav},\MSigma_k^\textrm{cav})$ \\
\bottomrule
\end{tabularx}

\newpage

\section{Gaussian Filtering} \label{app:kf_update}
Given observation model $p(\vy_k \mid \vf_k)=\N(\vy_k \mid \vf_k, \MSigma_k)$ for $\vf_k=\MH_k \vx_k$, along with current filter predictions \mbox{$p(\vx_k \mid \vy_{1:k-1}) = \N(\vx_k \mid \vm_k^{\mathrm{pred}}, \MP_k^{\mathrm{pred}})$}, the Kalman filter update equations are,
\begin{gather}
\begin{aligned} \label{eq:kf-update}
&\vmu_k = \MH_k \vm_k^{\mathrm{pred}}, \\
&\MS_k = \MH_k\MP_k^{\mathrm{pred}}\MH_k^\T + \MSigma_k, \\
&\MC_k = \MP_k^{\mathrm{pred}} \MH^\T, \\
&\MK_k = \MC_k \MS_k^{-1}, \\
&\vm_k = \vm_k^{\mathrm{pred}} + \MK_k (\vy_k - \vmu_k), \\
& \MP_k = \MP_k^{\mathrm{pred}} - \MK_k \MS_k \MK_k^\T.
\end{aligned}
\end{gather}
For nonlinear measurement models, $\vy_k = \vh(\vf_k, \vsigma_k)$, letting $\vmu_k^{\textrm{cav}} = \MH_k \vm_k^{\textrm{pred}}$ and $\MSigma_k^{\textrm{cav}} = \MH_k \MP_k^{\textrm{pred}} \MH_k^\T$, the statistical linear regression equations for the general Gaussian filtering methods are,
\begin{gather} \label{eq:gauss-filt-components}
\begin{aligned}
\vmu_k = \iint & \vh(\vf_k, \vsigma_k) \\
& \times \N(\vf_k \mid \vmu_k^{\mathrm{cav}}, \MSigma_k^{\mathrm{cav}}) \N(\vsigma_k \mid \bm{0}, \MSigma_k) \, \textrm{d}\vf_k \,\textrm{d}\vsigma_k, \\
\MS_k = \iint & (\vh(\vf_k, \vsigma_k) - \vmu_k) (\vh(\vf_k, \vsigma_k) - \vmu_k)^\T \\
& \times \N(\vf_k \mid \vmu_k^{\mathrm{cav}}, \MSigma_k^{\mathrm{cav}}) \N(\vsigma_k \mid \bm{0}, \MSigma_k) \, \textrm{d}\vf_k \,\textrm{d}\vsigma_k, \\
\MC_k = \iint & (\vf_k - \vmu_k^{\mathrm{cav}}) (\vh(\vf_k, \vsigma_k) - \vmu_k)^\T \\
& \times \N(\vf_k \mid \vmu_k^{\mathrm{cav}}, \MSigma_k^{\mathrm{cav}}) \N(\vsigma_k \mid \bm{0}, \MSigma_k) \, \textrm{d}\vf_k \,\textrm{d}\vsigma_k.
\end{aligned}
\end{gather}
Note that in the additive noise case, $\vh(\vf_k, \vsigma_k) = \tilde{\vh}(\vf_k) + \vsigma_k$, these can be simplified to,
\begin{gather} \label{eq:gauss-filt-components-additive}
\begin{aligned}
\vmu_k = \int & \tilde{\vh}(\vf_k) \N(\vf_k \mid \vmu_k^{\mathrm{cav}}, \MSigma_k^{\mathrm{cav}}) \, \textrm{d}\vf_k , \\
\MS_k = \int & \left[ (\tilde{\vh}(\vf_k) - \vmu_k) (\tilde{\vh}(\vf_k) - \vmu_k)^\T + \Cov[\vy_k \mid \vf_k] \right] \\
& \times \N(\vf_k \mid \vmu_k^{\mathrm{cav}}, \MSigma_k^{\mathrm{cav}})  \, \textrm{d}\vf_k , \\
\MC_k = \int & (\vf_k - \vmu_k^{\mathrm{cav}}) (\tilde{\vh}(\vf_k) - \vmu_k)^\T \N(\vf_k \mid \vmu_k^{\mathrm{cav}}, \MSigma_k^{\mathrm{cav}}) \, \textrm{d}\vf_k.
\end{aligned}
\end{gather}
for $\vsigma_k \sim \N(\bm{0}, \MSigma_k=\Cov[\vy_k \mid \vf_k])$. Note that we include the case where $\MSigma_k$ is a nonlinear function of $\vf_k$, which occurs in our approximations to discrete likelihoods presented in \cref{sec:app-models}. Here we have used $\tilde{\vh}(\vf_k)=\E[\vy_k \mid \vf_k]$.

\section{Closed-form Site Updates in \cref{sec:lin}}\label{app:site-updates}
Here we derive in full the closed form site updates after analytical linearisation in \cref{sec:lin}. Plugging the derivatives from \cref{eq:moment-match-derv} into the updates in \cref{eq:moment-match-filter} we get,
\begin{gather}
\begin{aligned} \label{eq:param-update-linear-app}
&\vmu_k^{\mathrm{site}} = \vmu_k^{\mathrm{cav}} + \left(\MJ_{\vf_k}^\T \hat{\MSigma}_k^{-1} \MJ_{\vf_k}\right)^{-1}  \MJ_{\vf_k}^\T \hat{\MSigma}_k^{-1} \vv_k , \\
&\MSigma_k^{\mathrm{site}} =   - \alpha\MSigma_k^{\mathrm{cav}} + \left(\MJ_{\vf_k}^\T \hat{\MSigma}_k^{-1} \MJ_{\vf_k}\right)^{-1} ,
\end{aligned}
\end{gather}
where $\vv_k=\vy_k-\vh(\vmu_k^{\mathrm{cav}},\bm{0})$. By the matrix inversion lemma, and letting $\MR_k = \MJ_{\vsigma_k} \MSigma_k \MJ^\T_{\vsigma_k}$,
\begin{multline} \label{eq:matrix-inv-lemma-app}
\hat{\MSigma}_k^{-1}= \MR_k^{-1} - \\
  \MR_k^{-1} \MJ_{\vf_k} \left( (\alpha\MSigma_k^{\mathrm{cav}})^{-1} + \MJ_{\vf_k}^\T \MR_k^{-1} \MJ_{\vf_k} \right)^{-1} \MJ_{\vf_k}^\T \MR_k^{-1},
\end{multline}
so that
\begin{equation} \label{eq:j-sigma-j-app}
\MJ_{\vf_k}^\T \hat{\MSigma}_k^{-1} \MJ_{\vf_k} = \MW_k -  \MW_k \left( (\alpha\MSigma_k^{\mathrm{cav}})^{-1} + \MW_k \right)^{-1} \MW_k,
\end{equation}
where $\MW_k = \MJ_{\vf_k}^\T \MR_k^{-1} \MJ_{\vf_k}$. Applying the matrix inversion lemma for a second time we obtain
\begin{align} \label{eq:d2lZ-app}
&\left(\MJ_{\vf_k}^\T \hat{\MSigma}_k^{-1} \MJ_{\vf_k}\right)^{-1} \nonumber \\
&= \MW_k^{-1} - \MW_k^{-1} \MW_k \bigg( \MW_k \MW_k^{-1} \MW_k \nonumber\\
& \hspace{9em} - \left( (\alpha\MSigma_k^{\mathrm{cav}})^{-1} + \MW_k \right) \bigg)^{-1} \MW_k \MW_k^{-1} \nonumber \\
&= \MW_k^{-1} + \alpha\MSigma_k^{\mathrm{cav}} \nonumber \\
&= \left( \MJ_{\vf_k}^\T \MR_k^{-1} \MJ_{\vf_k} \right)^{-1} + \alpha\MSigma_k^{\mathrm{cav}}. 
\end{align}
We can also write
\begin{multline} 
\hspace{-1.5em} \left(\MJ_{\vf_k}^\T \hat{\MSigma}_k^{-1} \MJ_{\vf_k}\right)^{-1} \MJ_{\vf_k}^\T \hat{\MSigma}_k^{-1} = \left( \left( \MJ_{\vf_k}^\T \MR_k^{-1} \MJ_{\vf_k} \right)^{-1} + \alpha \MSigma_k^{\mathrm{cav}} \right) \\ 
 \qquad\qquad \times \MJ_{\vf_k}^\T  \left(\MR_k + \alpha \MJ_{\vf_k} \MSigma_k^{\mathrm{cav}} \MJ_{\vf_k}^\T \right)^{-1}.
\end{multline}
Together the above calculations give the approximate site mean and covariance as
\begin{gather}
\begin{aligned} \label{eq:site-update-app}
&\MSigma_k^{\mathrm{site}} = \left( \MJ_{\vf_k}^\T \MR_k^{-1} \MJ_{\vf_k} \right)^{-1} , \\
&\vmu_k^{\mathrm{site}} = \vmu_k^{\mathrm{cav}} + \\
& \,\,\, \left( \MSigma_k^{\mathrm{site}} + \alpha \MSigma_k^{\mathrm{cav}} \right) \MJ_{\vf_k}^\T  \left(\MR_k + \alpha \MJ_{\vf_k} \MSigma_k^{\mathrm{cav}} \MJ_{\vf_k}^\T \right)^{-1}  \vv_k .
\end{aligned}
\end{gather}

\newpage 

\section{Analytical Linearisation in EP ($\alpha=1$) Results in an Iterated Version of the EKF}\label{app:derive-ekf}
Here we prove the result given in \cref{sec:lin}: a single pass of the proposed EP-style algorithm with analytical linearisation (\ie a first order Taylor series approximation) is exactly equivalent to the EKF. Plugging the closed form site updates, \cref{eq:site-update}, with $\alpha=1$ (since the filter predictions can be interpreted as the cavity with the \emph{full} site removed), into our modified Kalman filter update equations, \cref{eq:gaussian-update}, we get a new set of Kalman updates in which the latent noise terms are determined by scaling the observation noise with the Jacobian of the state. Crucially, on the first forward pass the Kalman prediction is used as the cavity such that $\MSigma_k^{\mathrm{cav}}=\MH_k\MP_k^{\mathrm{pred}}\MH_k^\T$:
\begin{gather}
\begin{aligned} \label{eq:lin-update-app}
&\MS_k = \MSigma_k^{\mathrm{cav}} + \left( \MJ_{\vf_k}^\T \MR_k^{-1} \MJ_{\vf_k} \right)^{-1}  ,\\
&\MK_k = \MP_k^{\mathrm{pred}} \MH_k^\T \MS_k^{-1}, \\
&\vm_k = \vm_k^{\mathrm{pred}} + \MK_k \MS_k \MJ_{\vf_k}^\T \left(\MR_k + \MJ_{\vf_k} \MSigma_k^{\mathrm{cav}} \MJ_{\vf_k}^\T \right)^{-1}  \vv_k , \\
&\MP_k = \MP_k^{\mathrm{pred}} - \MK_k \MS_k \MK_k^\T. \\
\end{aligned}
\end{gather}
where $\MR_k = \MJ_{\vsigma_k} \MSigma_k \MJ^\T_{\vsigma_k}$. This can be rewritten to explicitly show that there are two innovation covariance terms, $\MS_k$ and $\hat{\MS}_k$, which act on the state mean and covariance separately:

\paragraph{Linearised update step:}
\begin{gather}
\begin{aligned} \label{eq:lin-update-tidy-app}
&\hat{\MS}_k = \MSigma_k^{\mathrm{cav}} + \left( \MJ_{\vf_k}^\T \MR_k^{-1} \MJ_{\vf_k} \right)^{-1}  ,\\
&\MS_k = \MJ_{\vf_k} \MSigma_k^{\mathrm{cav}} \MJ_{\vf_k}^\T + \MR_k  ,\\
&\hat{\MK}_k = \MP_k^{\mathrm{pred}} \MH_k \hat{\MS}_k^{-1}, \\
&\MK_k = \MP_k^{\mathrm{pred}}\MH_k^\T \MJ_{\vf_k}^\T \MS_k^{-1}, \\
&\vm_k = \vm_k^{\mathrm{pred}} + \MK_k \vv_k    , \\
&\MP_k = \MP_k^{\mathrm{pred}} - \hat{\MK}_k \hat{\MS}_k \hat{\MK}_k^\T. \\
\end{aligned}
\end{gather}
Now we calculate the inverse of $\hat{\MS}_k$:
\begin{align} \label{eq:inv-S-hat-app}
\hat{\MS}_k^{-1} &= \left(\MSigma_k^{\mathrm{cav}} + \left( \MJ_{\vf_k}^\T \MR_k^{-1} \MJ_{\vf_k} \right)^{-1}\right)^{-1} \nonumber \\
&= \MJ_{\vf_k}^\T \MR_k^{-1} \MJ_{\vf_k}  - \nonumber \\
&\quad \MJ_{\vf_k}^\T \MR_k^{-1} \MJ_{\vf_k} \left( {\MSigma_k^{\mathrm{cav}}}^{-1} + \MJ_{\vf_k}^\T \MR_k^{-1} \MJ_{\vf_k} \right)^{-1} \MJ_{\vf_k}^\T \MR_k^{-1} \MJ_{\vf_k}
\end{align}
and the inverse of $\MS_k$:
\begin{align} \label{eq:inv-S-app}
\MS_k^{-1} &= \left( \MJ_{\vf_k} \MSigma_k^{\mathrm{cav}} \MJ_{\vf_k}^\T + \MR_k \right)^{-1} \nonumber \\
&= \MR_k^{-1} - \nonumber \\
& \quad \quad \MR_k^{-1} \MJ_{\vf_k} \left( {\MSigma_k^{\mathrm{cav}}}^{-1} + \MJ_{\vf_k}^\T \MR_k^{-1} \MJ_{\vf_k} \right)^{-1} \MJ_{\vf_k}^\T \MR_k^{-1} 
\end{align}
which shows that
\begin{gather}
\begin{aligned} \label{eq:inv-S-equivalence-app}
\hat{\MS}_k^{-1} &= \MJ_{\vf_k}^\T \MS_k^{-1} \MJ_{\vf_k},  \\
\end{aligned}
\end{gather}
and hence, recalling that $\MR_k = \MJ_{\vr_k} \MSigma_k \MJ^\T_{\vr_k}$, \cref{eq:lin-update-tidy-app} simplifies to give exactly the extended Kalman filter updates:

\paragraph{EKF update step:}
\begin{gather}
\begin{aligned} \label{eq:ekf-update-app}
&\MS_k = \MJ_{\vf_k} \MH_k\MP_k^{\mathrm{pred}}\MH_k^\T \MJ_{\vf_k}^\T + \MJ_{\vsigma_k} \MSigma_k \MJ^\T_{\vsigma_k}  ,\\
&\MK_k = \MP_k^{\mathrm{pred}}\MH_k^\T \MJ_{\vf_k}^\T \MS_k^{-1}, \\
&\vm_k = \vm_k^{\mathrm{pred}} + \MK_k (\vy_k-\vh(\MH_k\vm_k^{\mathrm{pred}},\bm{0}))    , \\
&\MP_k = \MP_k^{\mathrm{pred}} - \MK_k \MS_k \MK_k^\T. \\
\end{aligned}
\end{gather}

\section{General Gaussian Filter Site Updates in \cref{sec:gauss-filt}}\label{app:sl-site-updates}
Here we derive in full the site updates after statistical linear regression in \cref{sec:gauss-filt}. The Gaussian likelihood approximation results in,
\begin{align} 
\mathcal{L}_k &= \log \E_{q_k^{\textrm{cav}}}\big[\N^\alpha(\vy_k \mid \vmu_k+\MC_k^\T (\MSigma^\textrm{cav}_k)^{-1}(\vf_k-\vmu^\textrm{cav}_k), \MR_k) \,\big] \nonumber \\
&= c + \log \N\big(\vy_k \mid \vmu_k , \alpha^{-1} \tilde{\MSigma}_k \big),
\label{eq:moment-match-sl}
\end{align}
where $q_k^{\textrm{cav}}=\N(\vf_k \mid \vmu_k^{\mathrm{cav}},\MSigma_k^{\mathrm{cav}})$, $\MR_k = \MS_k-\MC_k^\T (\MSigma^\textrm{cav}_k)^{-1} \MC_k$ and $\tilde{\MSigma}_k=\MR_k + \alpha \MC_k^\T (\MSigma_k^{\mathrm{cav}})^{-1} \MC_k$ for $\vmu_k$, $\MS_k$ and $\MC_k$ given in \cref{eq:gauss-filt-components} with $\vmu^\textrm{cav}_k=\MH_k\vm^\textrm{pred}_k$, $\MSigma^\textrm{cav}_k=\MH_k\MP^\textrm{pred}_k\MH_k^\T$. Taking the derivatives of this log-Gaussian w.r.t.\ the cavity mean, we get
\begin{gather}
\begin{aligned} \label{eq:moment-match-sl-derv}
\bm{\nabla}{\mathcal{L}}_k &= \diff{\mathcal{L}_k}{\vmu_k^{\mathrm{cav}}} = \alpha \MOmega_k^\T \tilde{\MSigma}_k^{-1}\vv_k , \\
\bm{\nabla}^2{\mathcal{L}_k} &= \diffIIvec{\mathcal{L}_k}{\vmu_k^{\mathrm{cav}}}{(\vmu_k^{\mathrm{cav}})^\T} = - \alpha \MOmega_k^\T \tilde{\MSigma}_k^{-1} \MOmega_k,
\end{aligned}
\end{gather}
where $\vv_k=\vy_k-\vmu_k$ and
\begin{gather}
\begin{aligned} \label{eq:MU}
\MOmega_k &= \diff{\vmu_k}{\vmu_k^{\mathrm{cav}}} \\
&= \ \iint \vh(\vf_k, \vsigma_k) (\MSigma_k^{\mathrm{cav}})^{-1}(\vf_k - \vmu_k^{\textrm{cav}}) \N(\vf_k \mid \vmu_k^{\mathrm{cav}}, \MSigma_k^{\mathrm{cav}}) \\
& \hspace{3em} \times \N(\vsigma_k \mid \bm{0}, \MSigma_k) \, \textrm{d}\vf_k \,\textrm{d}\vsigma_k.
\end{aligned}
\end{gather}
As in \cref{sec:lin}, to ensure consistency we have assumed here that the derivative of $\tilde{\MSigma}_k$ is zero, despite the fact that it depends on $\vmu_k^{\textrm{cav}}$.

Plugging the derivatives from \cref{eq:moment-match-sl-derv} into the updates in \cref{eq:moment-match-filter} we get,
\begin{gather}
\begin{aligned} \label{eq:param-update-sl-app}
&\vmu_k^{\mathrm{site}} = \vmu_k^{\mathrm{cav}} + \left(\MOmega_k^\T \tilde{\MSigma}_k^{-1} \MOmega_k\right)^{-1} \MOmega_k^\T \tilde{\MSigma}_k^{-1} \vv_k , \\
&\MSigma_k^{\mathrm{site}} =  - \alpha \MSigma_k^{\mathrm{cav}} + \left(\MOmega_k^\T \tilde{\MSigma}_k^{-1} \MOmega_k \right)^{-1} .
\end{aligned}
\end{gather}

\newpage

\section{Avoiding Numerical Issues When Computing the Cavity} \label{app:cavity}

Computing the cavity distribution in \cref{eq:moment-match-smoother} involves the subtraction of two PSD covariance matrices. The result is not guaranteed to be PSD and not guaranteed to be invertible, which can lead to numerical issues. If $\vf_k$ is one-dimensional, then no such issue occurs. In the higher-dimensional case issues can be avoided by discarding the cross-covariances such that \cref{eq:moment-match-smoother} involves only element-wise subtraction of scalars. If using cubature to perform moment matching / linearisation, then this results in a loss of accuracy. However, for the Taylor series approximations (EKF / EKS / EEP) the cross-covariances are discarded anyway.

An alternative approach which does not trade off accuracy is to instead compute the cavity by taking the product of the forward and backward filtering distributions, an approach known as two-filter smoothing \citep{Sarkka:2013}, and then include a fraction $(1-\alpha)$ of the local site. This method only involves \emph{products} of PSD matrices which is more numerically stable. We did not implement this approach here.

\section{Marginal Likelihood Calculation During Filtering} \label{app:marg-lik}

The marginal likelihood, $p(\vy\mid\vtheta)$, is used as an optimisation objective for hyperparameter learning. The marginal likelihood can be written as a product of conditional terms (dropping the dependence on $\vtheta$ for notational convenience),
\begin{equation}
p(\vy) = p(\vy_1)\,p(\vy_2 \mid \vy_1)\,p(\vy_3 \mid \vy_{1:2})\prod_{k=4}^n p(\vy_k \mid \vy_{1:k-1}).
\end{equation}
Each term can be computed via numerical integration during the Kalman filter by noticing that,
\begin{align}
\!\!p(\vy_k \mid \vy_{1:k-1})\, &{=} \int p(\vy_k \mid \vx_k,\vy_{1:k-1}) p(\vx_k \mid \vy_{1:k-1}) \,\dd\vx_k \nonumber \\
&{=} \int p(\vy_k \mid \vf_k = \MH\vx_k) p(\vx_k \mid \vy_{1:k-1}) \,\dd\vx_k.
\end{align} 
The first component in the integral is the likelihood, and the second term is the forward filter prediction.

The Taylor series methods (EKF / EKS / EEP) aim to avoid numerical integration, and hence use an alternative approximation to the marginal likelihood based on the linearised model, as shown in \cref{alg:ekf}.

\clearpage

\onecolumn
\section{The EEP Algorithm} \label{app:algo}

\begin{algorithm*}[h!]
	\caption{EEP: Extended Expectation Propagation, a globally iterated Extended Kalman filter with power EP-style updates such that linearisation is performed w.r.t. the cavity mean.}
	\label{alg:ekf}
	\begin{algorithmic}
		\STATE {\bfseries Input:} $\{t_k, \vy_k\}_{k=1}^n$, $\MA_k$, $\MQ_k$, $\MP_\infty$, $\MSigma_k$ \comm{data, discrete state space model and obs. noise} \\
		\quad \qquad $\:$ $\vh$, $\MH_k$, $\MJ_{\vf}$, $\MJ_{\vsigma}$, $\alpha$            \comm{measurement model, Jacobian and EP power}\\
		$\vm_0 \leftarrow \bm{0}$, $\,$ $\MP_0 \leftarrow \MP_\infty$, $\,$ $\ve_{1:n}=\bm{0}$ \comm{initial state} \\
		\WHILE[\comm{iterated EP-style loop}]{not converged}
		\FOR[\comm{forward pass (FILTERING)}]{$k=1$ {\bfseries to} $n$}
		\STATE $\vm_k \leftarrow \MA_k\,\vm_{k-1}$ \comm{predict mean}
		\STATE $\MP_k \leftarrow \MA_k\, \MP_{k-1}\, \MA_k^{\top}{+}\MQ_k$ \comm{predict covariance}
		\IF{has label $\vy_k$}
		\STATE $\MSigma_k^{\mathrm{cav}} \leftarrow \MH_k\MP_k\MH_k^\T $ \comm{predict = forward cavity}  \\
		\STATE $\vmu_k^{\mathrm{cav}} \leftarrow \MH_k\vm_k $ \\
		\STATE $\vv_k \leftarrow \vy_k-\vh(\vmu_k^{\mathrm{cav}},\bm{0})$ \comm{residual} \\
		\STATE $\MJ_{\vf_k} \leftarrow \MJ_{\vf} |_{\vmu_k^{\mathrm{cav}},\bm{0}}$; \quad $\MJ_{\vsigma_k} \leftarrow \MJ_{\vsigma} |_{\vmu_k^{\mathrm{cav}},\bm{0}}$ \comm{evaluate Jacobians} \\
		\IF{first iteration}
		\STATE $\MSigma_k^{\mathrm{site}} \leftarrow \left( \MJ_{\vf_k}^\T \left(\MJ_{\vsigma_k} \MSigma_k \MJ_{\vsigma_k}^\T\right)^{-1} \MJ_{\vf_k} \right)^{-1}$ \comm{match moments to get site covariance...} \\
		\STATE $\vmu_k^{\mathrm{site}}  \leftarrow \vmu_k^{\mathrm{cav}} + \left(  \MSigma_k^{\mathrm{site}} + \MSigma_k^{\mathrm{cav}} \right) \MJ_{\vf_k}^\T  \left(\MJ_{\vsigma_k} \MSigma_k \MJ_{\vsigma_k}^\T+ \MJ_{\vf_k} \MSigma_k^{\mathrm{cav}} \MJ_{\vf_k}^\T \right)^{-1} \vv_k$ \comm{and site mean ($\alpha=1$)} \\
		\ENDIF
		\STATE $\MS_k \leftarrow \MH_k\MP_k\MH_k^\T + \MSigma_k^{\mathrm{site}}$ \comm{innovation} \\
		\STATE $\MK_k \leftarrow \MP_k\MH_k^\T \MS_k^{-1}$ \comm{Kalman gain} \\
		\STATE $\vm_{k} \leftarrow \vm_k + \MK_k (\vmu_k^{\mathrm{site}} - \vmu_k^{\mathrm{cav}})$ \comm{update mean} 
		\STATE $\MP_{k} \leftarrow \MP_k - \MK_k \MS_k \MK_k^\T$ \comm{update covariance}
		\STATE $\ME_k \leftarrow \MJ_{\vsigma_k} \MSigma_k \MJ_{\vsigma_k}^\T + \MJ_{\vf_k} \MSigma_k^{\mathrm{cav}} \MJ_{\vf_k}^\T$
		\STATE $\ve_k \leftarrow \frac{1}{2}  \log | 2\pi \ME_k | + \frac{1}{2}\vv_k^\T \ME_k^{-1} \vv_k$ \comm{energy} \\
		\ENDIF
		\ENDFOR
		\FOR[\comm{backward pass (SMOOTHING)}]{$k=n-1$ {\bfseries to} $1$}
		\STATE $\MG_k \leftarrow \MP_{k} \, \MA_{k+1}^\T \, (\MA_{k+1}\,\MP_{k}\,\MA_{k+1}^\T + \MQ_{k+1})^{-1}$ \comm{smoothing gain} \\
		\STATE $\vm_{k} \leftarrow \vm_{k} + \MG_k \, (\vm_{k+1} - \MA_{k+1}\,\vm_{k})$ \comm{update} \\
		\STATE $\MP_{k} \leftarrow \MP_{k} + \MG_k \, (\MP_{k+1} - \MA_{k+1}\,\MP_{k}\,\MA_{k+1}^\T - \MQ_{k+1}) \, \MG_k^\T$
		\IF{has label $\vy_k$}
		\STATE $\MSigma_k^{\mathrm{cav}} \leftarrow  \left( (\MH_k\MP_k\MH_k^\T)^{-1} - \alpha {\left(\MSigma_k^{\mathrm{site}}\right)}^{-1} \right)^{-1} $ \comm{remove site to get cavity covariance...} \\
		\STATE $\vmu_k^{\mathrm{cav}} \leftarrow \MSigma_k^{\mathrm{cav}} \left( (\MH_k\MP_k\MH_k^\T)^{-1} \MH_k\vm_k - \alpha {\left(\MSigma_k^{\mathrm{site}}\right)}^{-1} \vmu_k^{\mathrm{site}} \right)$ \comm{and cavity mean} \\
		\STATE $\MJ_{\vf_k} \leftarrow \MJ_{\vf} |_{\vmu_k^{\mathrm{cav}},\bm{0}}$; \quad $\MJ_{\vsigma_k} \leftarrow \MJ_{\vsigma} |_{\vmu_k^{\mathrm{cav}},\bm{0}}$ \comm{evaluate Jacobians} \\
		\STATE $\vv_k \leftarrow \vy_k-\vh(\vmu_k^{\mathrm{cav}},\bm{0})$ \comm{residual} \\
		\STATE $\MSigma_k^{\mathrm{site}} \leftarrow \left( \MJ_{\vf_k}^\T \left(\MJ_{\vsigma_k} \MSigma_k \MJ_{\vsigma_k}^\T\right)^{-1} \MJ_{\vf_k} \right)^{-1}$ \comm{match moments to get site covariance...} \\
		\STATE $\vmu_k^{\mathrm{site}}  \leftarrow \vmu_k^{\mathrm{cav}} + \left( \MSigma_k^{\mathrm{site}} + \alpha\MSigma_k^{\mathrm{cav}} \right) \MJ_{\vf_k}^\T  \left( \MJ_{\vsigma_k} \MSigma_k \MJ_{\vsigma_k}^\T + \alpha\MJ_{\vf_k} \MSigma_k^{\mathrm{cav}} \MJ_{\vf_k}^\T \right)^{-1} \vv_k$ \comm{and site mean}
		\ENDIF
		\ENDFOR   
		\ENDWHILE \\
		\STATE $\text{\bf Return:}~\mathbb{E}[\vf(t_k)] = \MH_k\vm_k; \, \mathbb{V}[\vf(t_k)] = \MH_k\MP_k \MH_k^{\T}$ \comm{posterior marginal mean and variance}
		\STATE $\phantom{\text{\bf Return:}}~\log p(\vy\mid\vtheta) \simeq -\sum_{k=1}^n \ve_k$ \comm{log marginal likelihood}
	\end{algorithmic}
\end{algorithm*}

\clearpage

\twocolumn

\newpage 

\section{Continuous Measurement Model Approximations for \cref{sec:results}} \label{sec:app-models}
The next subsections provide further details of the model formulations used in the experiments (\ie, how to write down approximative measurement models for common tasks such as heteroscedastic noise modelling, Poisson likelihoods, or logistic classification).

\subsection{Heteroscedastic Noise Model} 
The heteroscedastic noise model contains one GP for the mean, $f^{(1)}$, and another for the time-varying observation noise, $f^{(2)}$, both with Matern-$\nicefrac{3}{2}$ covariance functions. The GP priors are independent,
\begin{gather}
\begin{aligned}
    &f^{(1)}(t) \sim \GP(0, \kappa(t,t')), \\
    &f^{(2)}(t) \sim \GP(0, \kappa(t,t')),
\end{aligned}
\end{gather}
and the likelihood model is
\begin{equation}	
	\vy \mid \vf^{(1)},\vf^{(2)} \sim \prod_{k=1}^{n} \N(y_{k} \mid f^{(1)}_{k},[\phi(f^{(2)}_k)]^2).
\end{equation}
The corresponding state space observation model is
\begin{equation}
\vh(\vf_k, \sigma_k) = f^{(1)}_k + \phi(f^{(2)}_k) \sigma_k,
\end{equation}
where $\sigma_k\sim\N(0,1)$ and $\phi(f) = \log(1+\exp(f-\frac{1}{2}))$. 
The Jacobians w.r.t.\ the (two-dimensional) latent GPs $\vf_k$ and the noise variable $\sigma_k$ are,
\begin{subequations}
  \begin{align}
    &\MJ_{\vf}(\vf_k, \sigma_k) 
      = \frac{\partial \bar{\vh}}{\partial \vf_k^\T} =  \left[ 1, \, \phi ' \big(f^{(2)}_k\big) \sigma_k \right], \\
	&\MJ_{\vsigma}(\vf_k, \sigma_k) = \frac{\partial \bar{\vh}}{\partial \sigma_k^\T} = \phi \big(f^{(2)}_k\big),
  \end{align}
\end{subequations}
where the derivative of the softplus is the sigmoid function:
\begin{equation}
\phi'(f) = \frac{1}{1+\exp(-f+\frac{1}{2})}.
\end{equation}
In practice a problem arises when using the above linearisation. Since the mean of $\sigma_k$ is zero, the Jacobian w.r.t.\ $f^{(2)}$ disappears when evaluated at the mean regardless of the value of $f^{(2)}$. This means that the second latent function is never updated, which results in poor performance, as shown in \cref{tbl:results}. We found that statistical linearisation suffers from a similar issue, providing little importance to the latent function that models the noise, which highlights a potential weakness of linearisation-based methods.

\begin{figure*}[h!]
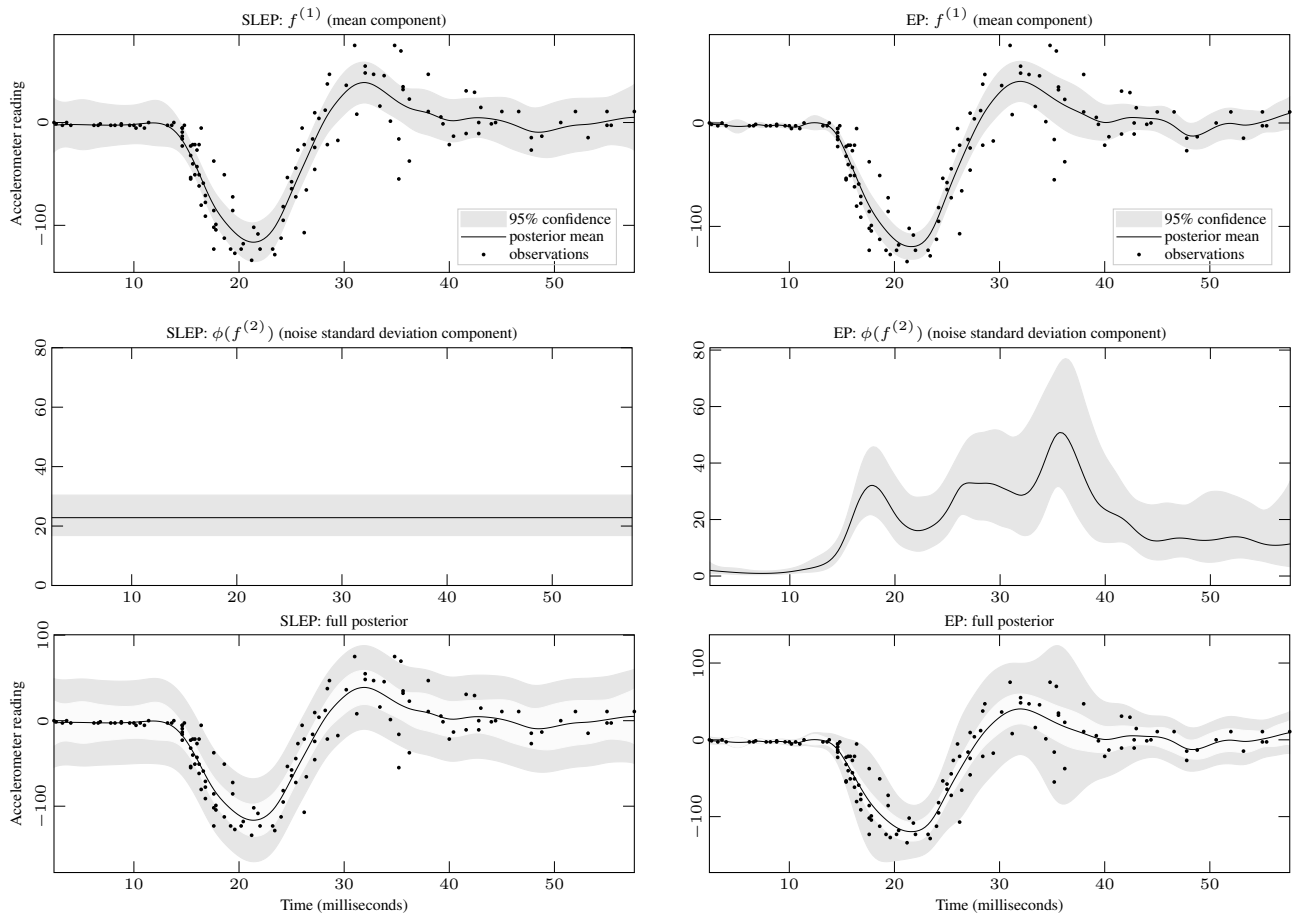

	\centering\tiny
	\pgfplotsset{yticklabel style={rotate=90}, ylabel style={yshift=0pt},scale only axis,axis on top,clip=false,title style={yshift=-6pt}}
	\setlength{\figurewidth}{.45\textwidth}
	\setlength{\figureheight}{.41\figurewidth}
	\begin{subfigure}[t]{.49\textwidth}
		\raggedright
		\input{./fig/mcycle_slep_f1.tex}
	\end{subfigure}
	\hspace*{\fill}
	\begin{subfigure}[t]{.49\textwidth}
		\raggedleft
		\input{./fig/mcycle_f1.tex} 
	\end{subfigure}\\
	\pgfplotsset{ignore legend}
	\begin{subfigure}[t]{.49\textwidth}
		\raggedright
		\input{./fig/mcycle_slep_f2.tex}
	\end{subfigure}
	\hspace*{\fill}
	\begin{subfigure}[t]{.49\textwidth}
		\raggedleft
		\input{./fig/mcycle_f2.tex}
	\end{subfigure}\\
	\begin{subfigure}[t]{.49\textwidth}
		\raggedright
		\input{./fig/mcycle_slep_combined.tex}
	\end{subfigure}
	\hspace*{\fill}
	\begin{subfigure}[t]{.49\textwidth}
		\raggedleft
		\input{./fig/mcycle_combined.tex}
	\end{subfigure}\\
	\vspace{-0.2cm}
	\caption{Model components for the motorcycle crash experiment. \textbf{Left} is the SLEP method (with Gauss-Hermite cubature, \ie GHEP) and \textbf{right} is the EP equivalent. The linearisation-based methods fail to incorporate the heteroscedastic noise, whereas EP captures rich time-varying behaviour. The \textbf{top} plots are the posterior for $f^{(1)}(t)$ (the mean process), the \textbf{middle} plots show the posterior for $f^{(2)}(t)$ (the observation noise process), and the \textbf{bottom} plots show the full model.} 
	\label{fig:full_mcycle}
\end{figure*}

\cref{fig:full_mcycle} plots a breakdown of the different components in the posterior for the motorcycle crash data set.

\subsection{Log-Gaussian Cox Process} 
For a log-Gaussian Cox process, binning the data into subregions and assuming the process has locally constant intensity in these subregions allows us to use a Poisson likelihood, $p(\vy \mid \vf) \approx \prod_{k=1}^n \mathrm{Poisson}(\vy_k \mid \exp(\vf(\hat{t}_k)))$, where $\hat{t}_k$ is the bin coordinate and $\vy_k$ the number of data points in it. However, the Poisson is a discrete probability distribution and the EKF and EEP methods requires the observation model to be differentiable. Therefore we use a Gaussian approximation, noticing that the first two moments of the Poisson distribution are equal to the intensity $\vlambda_k=\exp\big(\vf_k\big)$.

We have a GP prior over $\vf$:
\begin{equation}
\vf(t)  \sim \GP(\bm{0}, \MK(t,t')),
\end{equation}
and the approximate Gaussian likelihood is
\begin{equation}
p(\vy \mid \vf) = \prod_{k=1}^{n} \N\left(\vy_k \mid \exp(\vf_k),\, \textrm{diag}\!\left[\exp(\vf_k)\right] \right),
\end{equation}
where $\textrm{diag}\!\left[\exp(\vf_k)\right]$ is a diagonal matrix whose entries are the elements of $\exp(\vf_k)$. This implies the following state space observation model:
\begin{equation}
\vh(\vf_k, \vsigma_k) = \exp(\vf_k)+ \textrm{diag}\!\left[\exp(\vf_k/2)\right] \vsigma_k,
\end{equation}
where $\vsigma_k \sim \N(\bm{0}, \MI)$. The EKF and EEP algorithms require the Jacobian of $\vh(\vf_k,\vsigma_k)$ with respect to $\vf_k$ and $\vsigma_k$, which are given by,
\begin{subequations}
  \begin{align} \label{eq:coal}
  &\MJ_{\vf}(\vf_k, \vsigma_k) = \frac{\partial \vh}{\partial \vf_k^\T} \nonumber \\
  &= \textrm{diag}\!\left[\exp(\vf_k)+ \frac{1}{2}\textrm{diag}[\exp(\vf_k/2)]\vsigma_k\right], \\
  &\MJ_{\vsigma}(\vf_k, \vsigma_k) = \frac{\partial \vh}{\partial \vsigma_k^\T} =  \textrm{diag}\!\left[\exp(\vf_k/2)\right].
  \end{align}
\end{subequations}

\subsection{Bernoulli (Logistic Classification)} 
As in standard GP classification we place a GP prior over the latent function $f$, and use a Bernoulli likelihood by mapping $f$ through the logistic function $\psi(f)= \frac{1}{1+ \exp(-f)}$,
\begin{subequations}
\begin{align} 
& f(t)\sim \GP(0, \kappa(t,t')), \\  
& \vy \mid \vf \sim \prod_{k=1}^{n}  \mathrm{Bern} (\psi(f(t_k))).
\end{align}
\end{subequations}
As with the Poisson likelihood, we wish to approximate the Bernoulli with a distribution that has continuous support. We form a Gaussian approximation whose mean and variance are equal to that of the Bernoulli distribution, which has mean $\E[y\mid f]=\psi(f)$, and variance $\Var[y \mid f]=\psi(f)(1-\psi(f))$, giving:
\begin{equation}
\vy \mid \vf \sim \prod_{k=1}^{n} \N\big(\,y_k \mid \psi(f_k), \, \psi(f_k)(1-\psi(f_k)) \, \big).
\end{equation}
Therefore the approximate state space observation model is
\begin{equation}
  h(f_k,\sigma_k) = \frac{1}{1+ \exp(-f_k)} + \frac{\exp(f_k/2)}{1 + \exp(f_k)} \sigma_k,
\end{equation}
and the Jacobians are
\begin{subequations}
\begin{align}
&\MJ_{f}(f_k,\sigma_k) = \frac{\partial h}{\partial f_k} \nonumber \\
&= \frac{\exp(f_k)}{(1+\exp(f_k))^2} +  \frac{\exp(f_k/2) - \exp(3f_k/2)}{2(1+\exp(f))^2}  \sigma_k, \\
&\MJ_{\sigma}(f_k,\sigma_k) = \frac{\partial h}{\partial \sigma_k} = \frac{\exp(f_k/2)}{1+\exp(f_k)}.
\end{align}
\end{subequations}

\subsection{Bernoulli (Probit Classification)} 

The Probit likelihood can be constructed similarly to the Logistic model above, by simply swapping the logistic function for the Normal CDF: $\psi(f)= \Phi(f) = \int_{-\infty}^f \N(x \mid 0, 1) \dd x$.

\medskip


\section{Supplementary Figures for \cref{sec:results}}

\subsection{Rainforest}

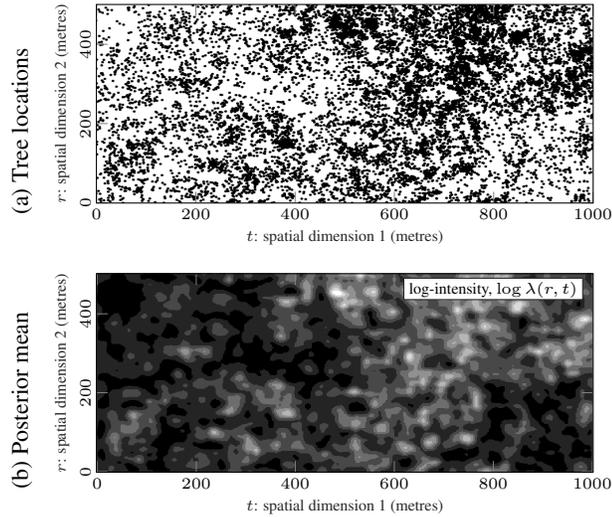
\begin{figure}[h!]
	\centering\tiny
	\newcommand{\mycaption}[1]{\rotatebox{90}{\parbox[c][2em][c]{1.2\figureheight}{\centering\footnotesize #1}}}
	\pgfplotsset{yticklabel style={rotate=90}, ylabel style={yshift=0pt},scale only axis,axis on top,clip=false}
	\setlength{\figurewidth}{.8\columnwidth}
	\setlength{\figureheight}{.4\figurewidth}
	\begin{subfigure}[t]{\columnwidth}
		\raggedright
		\pgfplotsset{ignore legend}
		\mycaption{(a)~Tree locations}\phantomsubcaption~~
%
%
\begin{tikzpicture}

\begin{axis}[%
axis on top,
xmin=-0.275806451612903,
xmax=1000.47580645161,
xlabel style={font=\color{white!15!black}},
xlabel={$t$: spatial dimension 1 (metres)},
ymin=-0.375691244239631,
ymax=500.47569124424,
ylabel style={font=\color{white!15!black}},
ylabel={$r$: spatial dimension 2 (metres)},
axis background/.style={fill=white},
legend style={legend cell align=left, align=left, draw=white!15!black},
width=\figurewidth,
height=\figureheight
]
\addplot [forget plot] graphics [xmin=-0.275806451612903, xmax=1000.47580645161, ymin=-0.375691244239631, ymax=500.47569124424] {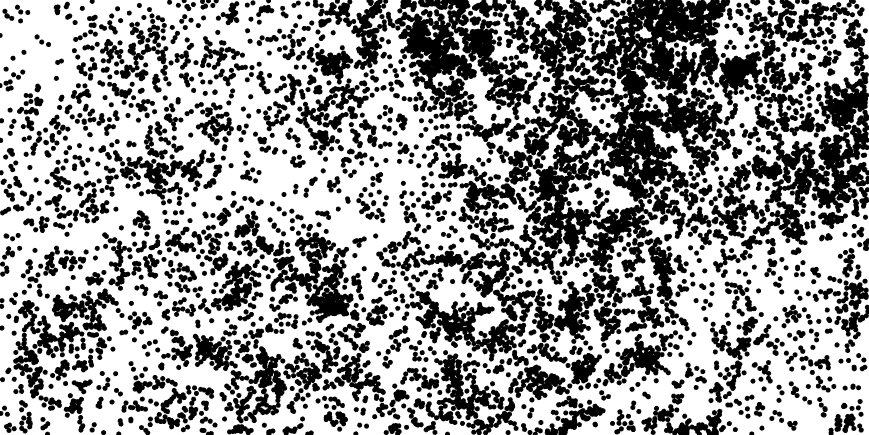};
\end{axis}
\end{tikzpicture}%
		\label{fig:rainforest-app1}  
	\end{subfigure}\\
\vspace{1em}
	\begin{subfigure}[t]{\columnwidth}
		\raggedright
		\mycaption{(b)~Posterior mean}\phantomsubcaption~~	  
%
%
\begin{tikzpicture}

\begin{axis}[%
axis on top,
xmin=-1.6410371763936,
xmax=1001.64103717639,
xlabel style={font=\color{white!15!black}},
xlabel={$t$: spatial dimension 1 (metres)},
ymin=-1.64433243176789,
ymax=501.644332431768,
ylabel style={font=\color{white!15!black}},
ylabel={$r$: spatial dimension 2 (metres)},
axis background/.style={fill=white},
legend style={legend cell align=left, align=left, draw=white!15!black,fill=white},
width=\figurewidth,
height=\figureheight,
legend image post style={scale=0}
]
\addlegendimage{only marks,no markers}
\addlegendentry{ log-intensity, $\log \lambda(r,t)$}
\addplot [forget plot] graphics [xmin=-1.6410371763936, xmax=1001.64103717639, ymin=-1.64433243176789, ymax=501.644332431768] {./fig/rainforest-1.png};

\end{axis}
\end{tikzpicture}%
		\label{fig:rainforest-app2}  
	\end{subfigure}
	\vspace*{-2em}
	\caption{The data (a) are $12{,}929$ tree locations in a rainforest. They are binned into a grid of $500 \times 250$ and we apply a log-Gaussian Cox process using EEP for inference. The posterior log-intensity is shown in (b).}
	\label{fig:rainforest-app}
	\vspace*{-2em}
\end{figure}

\newpage

\subsection{Audio}

\begin{figure}[h!]
	\centering\tiny
	\newcommand{\mycaption}[1]{\rotatebox{90}{\parbox[c][2em][c]{1.2\figureheight}{\centering\footnotesize #1}}}
	\pgfplotsset{yticklabel style={rotate=90}, ylabel style={yshift=0pt},scale only axis,axis on top,clip=false}
	\setlength{\figurewidth}{.8\columnwidth}
	\setlength{\figureheight}{.4\figurewidth}
	\begin{subfigure}[t]{\columnwidth}
		\raggedright
		\pgfplotsset{ignore legend}
		\mycaption{(a)~Audio signal}\phantomsubcaption~~
%
%
\begin{tikzpicture}

\begin{axis}[%
axis on top,
xmin=-0.000288350634371396,
xmax=0.500288350634371,
ymin=-1.00146198830409,
ymax=1.00146198830409,
ylabel style={font=\color{white!15!black}},
ylabel={$y$},
axis background/.style={fill=white},
legend style={legend cell align=left, align=left, draw=white!15!black},
width=\figurewidth,
height=\figureheight
]
\addplot [forget plot] graphics [xmin=-0.000288350634371396, xmax=0.500288350634371, ymin=-1.00146198830409, ymax=1.00146198830409] {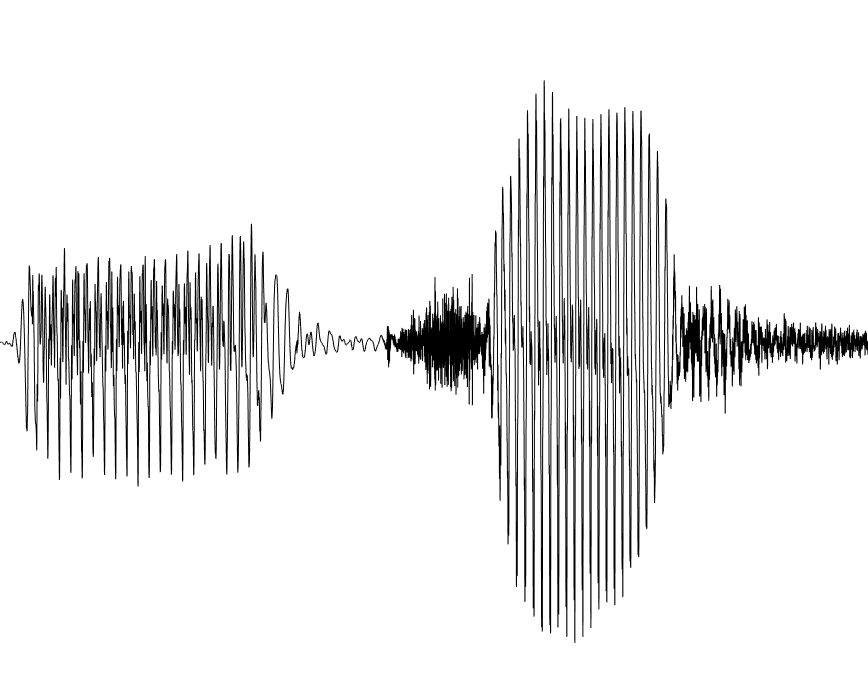};
\end{axis}
\end{tikzpicture}%
		\label{fig:audio-app1}  
	\end{subfigure}\\
	\vspace{1em}
	\begin{subfigure}[t]{\columnwidth}
		\raggedright
		\mycaption{(b)~First component}\phantomsubcaption~~	  
%
%
\begin{tikzpicture}

\begin{axis}[%
axis on top,
xmin=-0.000288350634371396,
xmax=0.500288350634371,
ymin=-0.5,
ymax=0.5,
ytick={-0.4,    0,  0.4},
ylabel style={font=\color{white!15!black}},
ylabel={$\phantom{y}$},
axis background/.style={fill=white},
legend style={legend cell align=left, align=left, draw=white!15!black},
width=\figurewidth,
height=\figureheight
]
\addplot [forget plot] graphics [xmin=-0.000288350634371396, xmax=0.500288350634371, ymin=-0.250365497076023, ymax=0.250365497076023] {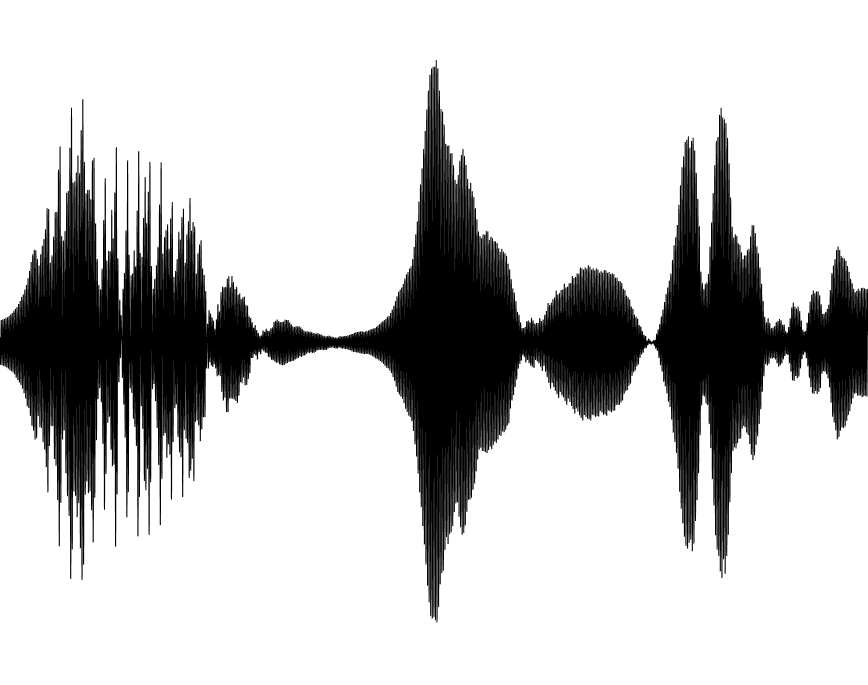};
\addplot [color=black]
  table[row sep=crcr]{%
0	0\\
};
\addlegendentry{periodic subband (posterior mean)}

\addplot [color=red, line width=1.0pt]
  table[row sep=crcr]{%
2.26757369614512e-05	0.0269148391009397\\
0.00342403628117914	0.0274388538342502\\
0.00682539682539683	0.0286723222998524\\
0.0102267573696145	0.0314002546496186\\
0.0136281179138322	0.0370930984597124\\
0.0170294784580499	0.0480007890615269\\
0.0204308390022676	0.0660389512782576\\
0.0238321995464853	0.0918185279392326\\
0.0272335600907029	0.125261261119925\\
0.0306349206349206	0.159689640956271\\
0.0340362811791383	0.188887807016909\\
0.037437641723356	0.203856016826185\\
0.0408390022675737	0.208161954311482\\
0.0442403628117914	0.203409667301508\\
0.0476417233560091	0.193959579979501\\
0.0510430839002268	0.183864868352142\\
0.0544444444444444	0.18552120289588\\
0.0578458049886621	0.201098554242049\\
0.0612471655328798	0.228712371275753\\
0.0646485260770975	0.254376854250199\\
0.0680498866213152	0.268379910753034\\
0.0714512471655329	0.270697389604722\\
0.0748526077097506	0.26900881575955\\
0.0782539682539683	0.266708379158464\\
0.0816553287981859	0.263997058138167\\
0.0850566893424036	0.256318990068171\\
0.0884580498866213	0.243726729057858\\
0.091859410430839	0.230937982289504\\
0.0952607709750567	0.216225660418717\\
0.0986621315192744	0.201436783477038\\
0.102063492063492	0.190459634144645\\
0.10546485260771	0.183095936220048\\
0.108866213151927	0.172991962859813\\
0.112267573696145	0.159282992598342\\
0.115668934240363	0.143126954762809\\
0.119070294784581	0.125743565910402\\
0.122471655328798	0.108330458771517\\
0.125873015873016	0.0947492826969674\\
0.129274376417234	0.0868752317855123\\
0.132675736961451	0.0833316444794808\\
0.136077097505669	0.0806448245810429\\
0.139478458049887	0.0754871289232049\\
0.142879818594104	0.0668685370380839\\
0.146281179138322	0.0567688079090762\\
0.14968253968254	0.0475717348745669\\
0.153083900226757	0.0404739207745575\\
0.156485260770975	0.0356354430271397\\
0.159886621315193	0.032531794194302\\
0.16328798185941	0.0305411323121336\\
0.166689342403628	0.0292333689268112\\
0.170090702947846	0.0283405416136953\\
0.173492063492064	0.0277321872901192\\
0.176893424036281	0.0273462458589964\\
0.180294784580499	0.0271295856462093\\
0.183696145124717	0.0270160739411457\\
0.187097505668934	0.0269475989820077\\
0.190498866213152	0.0268912813626562\\
0.19390022675737	0.0268308860516329\\
0.197301587301587	0.0267671853795797\\
0.200702947845805	0.0267174776899698\\
0.204104308390023	0.0266890497463333\\
0.20750566893424	0.0266701981535687\\
0.210907029478458	0.0266518093367385\\
0.214308390022676	0.0266306955849851\\
0.217709750566893	0.0266006322250266\\
0.221111111111111	0.0265601660579284\\
0.224512471655329	0.026527162013446\\
0.227913832199546	0.0265331194692187\\
0.231315192743764	0.0266057226503707\\
0.234716553287982	0.0270501368072902\\
0.2381179138322	0.0287492927324944\\
0.241519274376417	0.033065284464607\\
0.244920634920635	0.0401033014708592\\
0.248321995464853	0.0462996523181558\\
0.25172335600907	0.0464036438508675\\
0.255124716553288	0.0414229945237718\\
0.258526077097506	0.0371918822273576\\
0.261927437641723	0.0357416957425908\\
0.265328798185941	0.0373249510000061\\
0.268730158730159	0.0386955030154774\\
0.272131519274376	0.0376491661021246\\
0.275532879818594	0.0353786106722699\\
0.278934240362812	0.033710253611926\\
0.282335600907029	0.0329638192485063\\
0.285736961451247	0.0328491467583242\\
0.289138321995465	0.033219725449833\\
0.292539682539683	0.0339349578889612\\
0.2959410430839	0.0349272560024493\\
0.299342403628118	0.0366143202181858\\
0.302743764172336	0.0391458629799018\\
0.306145124716553	0.0422247010795161\\
0.309546485260771	0.0452878679563916\\
0.312947845804989	0.0477184356106751\\
0.316349206349206	0.0490272146045185\\
0.319750566893424	0.0490821854558704\\
0.323151927437642	0.0482446318814966\\
0.326553287981859	0.0469039637885901\\
0.329954648526077	0.0452839636511236\\
0.333356009070295	0.0435374412980336\\
0.336757369614512	0.0417059603542039\\
0.34015873015873	0.0398478635850647\\
0.343560090702948	0.0380627316113704\\
0.346961451247166	0.0364360273626979\\
0.350362811791383	0.0349640809514735\\
0.353764172335601	0.0335821318600092\\
0.357165532879819	0.0322664398204323\\
0.360566893424036	0.0310342032433428\\
0.363968253968254	0.030009494982598\\
0.367369614512472	0.0292994091899202\\
0.370770975056689	0.028911288042683\\
0.374172335600907	0.0288168054983968\\
0.377573696145125	0.0290453776494557\\
0.380975056689342	0.0297846543910169\\
0.38437641723356	0.0314976290563907\\
0.387777777777778	0.034952902069768\\
0.391179138321995	0.040881917263681\\
0.394580498866213	0.0482481273058267\\
0.397981859410431	0.0532389873642206\\
0.401383219954649	0.0530017854904407\\
0.404784580498866	0.0497966915605422\\
0.408185941043084	0.050855659780496\\
0.411587301587302	0.0582757045386519\\
0.414988662131519	0.0653130431592647\\
0.418390022675737	0.0633264093493943\\
0.421791383219955	0.0540619109220361\\
0.425192743764172	0.0467051919321924\\
0.42859410430839	0.0425352808362946\\
0.431995464852608	0.0396808276995232\\
0.435396825396825	0.0367986370747047\\
0.438798185941043	0.0335589839694244\\
0.442199546485261	0.0310356389168999\\
0.445600907029478	0.0296106457155368\\
0.449002267573696	0.029019465660993\\
0.452403628117914	0.0289309001274837\\
0.455804988662132	0.0291347653822276\\
0.459206349206349	0.0293377750194234\\
0.462607709750567	0.0295047911374145\\
0.466009070294785	0.0298254449038209\\
0.469410430839002	0.0301215883075897\\
0.47281179138322	0.0302353729582364\\
0.476213151927438	0.0305160343015117\\
0.479614512471655	0.0311346934915647\\
0.483015873015873	0.0311841896239905\\
0.486417233560091	0.03033436555155\\
0.489818594104308	0.0292516246100282\\
0.493219954648526	0.028409765685502\\
0.496621315192744	0.0278505156700028\\
};
\addlegendentry{positive amplitude (posterior mean)}

\end{axis}
\end{tikzpicture}%
		\label{fig:audio-app2}  
	\end{subfigure}
    \vspace{1em}
    \pgfplotsset{ignore legend}
    \begin{subfigure}[t]{\columnwidth}
    	\raggedright
    	\mycaption{(c)~Second component}\phantomsubcaption~~	  
%
%
\begin{tikzpicture}

\begin{axis}[%
axis on top,
xmin=-0.000288350634371396,
xmax=0.500288350634371,
ymin=-1,
ymax=1,
ylabel style={font=\color{white!15!black}},
ylabel={$\phantom{y}$},
axis background/.style={fill=white},
legend style={legend cell align=left, align=left, draw=white!15!black},
width=\figurewidth,
height=\figureheight
]
\addplot [forget plot] graphics [xmin=-0.000288350634371396, xmax=0.500288350634371, ymin=-0.500730994152047, ymax=0.500730994152047] {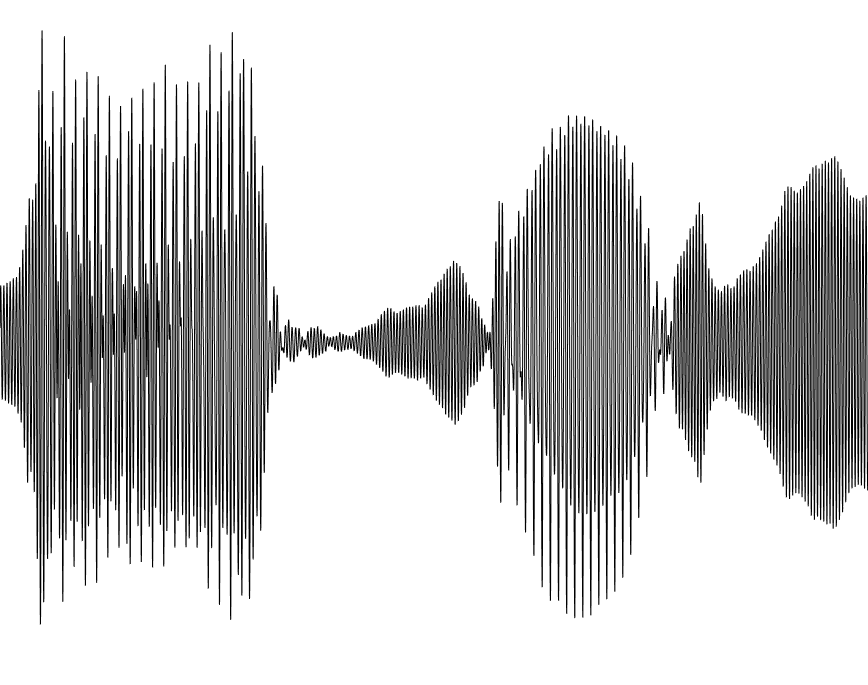};
\addplot [color=black]
  table[row sep=crcr]{%
0	0\\
};
\addlegendentry{data1}

\addplot [color=red, line width=1.0pt]
  table[row sep=crcr]{%
2.26757369614512e-05	0.0187195746855965\\
0.00342403628117914	0.0206404633239644\\
0.00682539682539683	0.0256824427633004\\
0.0102267573696145	0.0385157104434117\\
0.0136281179138322	0.0667088333579075\\
0.0170294784580499	0.111469217737306\\
0.0204308390022676	0.169616043181015\\
0.0238321995464853	0.23392767119555\\
0.0272335600907029	0.277867752414711\\
0.0306349206349206	0.295492994432641\\
0.0340362811791383	0.297316519400175\\
0.037437641723356	0.290194253080444\\
0.0408390022675737	0.277694832692301\\
0.0442403628117914	0.267840391836239\\
0.0476417233560091	0.263757732847097\\
0.0510430839002268	0.263873662214645\\
0.0544444444444444	0.265968613798119\\
0.0578458049886621	0.268176007096177\\
0.0612471655328798	0.269559380529394\\
0.0646485260770975	0.270457751521046\\
0.0680498866213152	0.269664280390202\\
0.0714512471655329	0.267057479637494\\
0.0748526077097506	0.264623387059667\\
0.0782539682539683	0.264370211103869\\
0.0816553287981859	0.26899089448583\\
0.0850566893424036	0.276018564318243\\
0.0884580498866213	0.284693306530034\\
0.091859410430839	0.293118277301597\\
0.0952607709750567	0.299994072599874\\
0.0986621315192744	0.303915249009615\\
0.102063492063492	0.308717364981188\\
0.10546485260771	0.31699112886779\\
0.108866213151927	0.330071635142004\\
0.112267573696145	0.345606379028021\\
0.115668934240363	0.361256093728664\\
0.119070294784581	0.37153630869925\\
0.122471655328798	0.372650379777492\\
0.125873015873016	0.370039393251781\\
0.129274376417234	0.369770250975193\\
0.132675736961451	0.37489289357059\\
0.136077097505669	0.377805667424403\\
0.139478458049887	0.369975862501534\\
0.142879818594104	0.34451285770823\\
0.146281179138322	0.300828751990381\\
0.14968253968254	0.242941305011391\\
0.153083900226757	0.18099734759957\\
0.156485260770975	0.126557211535194\\
0.159886621315193	0.0857538554293138\\
0.16328798185941	0.0584982724067091\\
0.166689342403628	0.0419002319391121\\
0.170090702947846	0.0319710580091139\\
0.173492063492064	0.0260017205613845\\
0.176893424036281	0.0225630708470164\\
0.180294784580499	0.0206683175259825\\
0.183696145124717	0.0195907829153593\\
0.187097505668934	0.0189337657437304\\
0.190498866213152	0.018540447755333\\
0.19390022675737	0.0183137455657702\\
0.197301587301587	0.0181429431301079\\
0.200702947845805	0.0180039952822531\\
0.204104308390023	0.0179447953849204\\
0.20750566893424	0.0179445963893352\\
0.210907029478458	0.0179489105273527\\
0.214308390022676	0.0179886305678056\\
0.217709750566893	0.0181420930047965\\
0.221111111111111	0.0183293650529628\\
0.224512471655329	0.0183066691207031\\
0.227913832199546	0.0180603443607304\\
0.231315192743764	0.0179117008103197\\
0.234716553287982	0.0178738254671082\\
0.2381179138322	0.0178081679650133\\
0.241519274376417	0.0177580923609763\\
0.244920634920635	0.0179938784814149\\
0.248321995464853	0.0187778681889142\\
0.25172335600907	0.0199062747470309\\
0.255124716553288	0.0212311202774165\\
0.258526077097506	0.022702239558386\\
0.261927437641723	0.0237750491270023\\
0.265328798185941	0.0238610419842974\\
0.268730158730159	0.0238111837803863\\
0.272131519274376	0.0257309763845619\\
0.275532879818594	0.0322328561260663\\
0.278934240362812	0.0474207522797749\\
0.282335600907029	0.0783648127433489\\
0.285736961451247	0.12831929043862\\
0.289138321995465	0.185162263353342\\
0.292539682539683	0.233214710634001\\
0.2959410430839	0.273362096457362\\
0.299342403628118	0.312287147064744\\
0.302743764172336	0.353055736510858\\
0.306145124716553	0.394784879034023\\
0.309546485260771	0.434886568022199\\
0.312947845804989	0.470994628727824\\
0.316349206349206	0.501700036590872\\
0.319750566893424	0.527642694815007\\
0.323151927437642	0.550207965960892\\
0.326553287981859	0.568918545851898\\
0.329954648526077	0.581469562853779\\
0.333356009070295	0.587021579580808\\
0.336757369614512	0.586071030137084\\
0.34015873015873	0.57935253091799\\
0.343560090702948	0.56809340409418\\
0.346961451247166	0.553794567565752\\
0.350362811791383	0.536492040907733\\
0.353764172335601	0.514842336661039\\
0.357165532879819	0.487307589097046\\
0.360566893424036	0.452365717866045\\
0.363968253968254	0.408439368683962\\
0.367369614512472	0.354678617816778\\
0.370770975056689	0.292732490867898\\
0.374172335600907	0.227849806833656\\
0.377573696145125	0.168233369895924\\
0.380975056689342	0.120430094127015\\
0.38437641723356	0.0860295409533935\\
0.387777777777778	0.0634642613673885\\
0.391179138321995	0.0496381208399254\\
0.394580498866213	0.0425741347598545\\
0.397981859410431	0.0409119087416283\\
0.401383219954649	0.0408491492755151\\
0.404784580498866	0.0360169179885822\\
0.408185941043084	0.0269965414728105\\
0.411587301587302	0.0211682030066475\\
0.414988662131519	0.0189535922675483\\
0.418390022675737	0.0185318805963123\\
0.421791383219955	0.0186922046272364\\
0.425192743764172	0.0192810431264042\\
0.42859410430839	0.0194606147102028\\
0.431995464852608	0.0188820866934734\\
0.435396825396825	0.018369685549402\\
0.438798185941043	0.0183595155311906\\
0.442199546485261	0.0187956917771402\\
0.445600907029478	0.0196411710813892\\
0.449002267573696	0.021467296445143\\
0.452403628117914	0.0237915108493008\\
0.455804988662132	0.0238812265655672\\
0.459206349206349	0.023000355424213\\
0.462607709750567	0.023969299306464\\
0.466009070294785	0.0262543405616685\\
0.469410430839002	0.0272545227991545\\
0.47281179138322	0.0262898977137067\\
0.476213151927438	0.0256398375097226\\
0.479614512471655	0.0254260302493765\\
0.483015873015873	0.0235303464315521\\
0.486417233560091	0.0196805170101133\\
0.489818594104308	0.0165684879450751\\
0.493219954648526	0.0156812013473797\\
0.496621315192744	0.0163870941202017\\
};
\addlegendentry{data2}

\end{axis}
\end{tikzpicture}%
    	\label{fig:audio-app3}  
    \end{subfigure}
    \vspace{1em}
    \begin{subfigure}[t]{\columnwidth}
    	\raggedright
    	\mycaption{(d)~Third component}\phantomsubcaption~~	  
%
%
\begin{tikzpicture}

\begin{axis}[%
axis on top,
xmin=-0.000288350634371396,
xmax=0.500288350634371,
xlabel style={font=\color{white!15!black}},
xlabel={$t$: time (seconds)},
ymin=-1,
ymax=1,
ylabel style={font=\color{white!15!black}},
ylabel={$\phantom{y}$},
axis background/.style={fill=white},
legend style={legend cell align=left, align=left, draw=white!15!black},
width=\figurewidth,
height=\figureheight
]
\addplot [forget plot] graphics [xmin=-0.000288350634371396, xmax=0.500288350634371, ymin=-0.801166180758017, ymax=0.801166180758017] {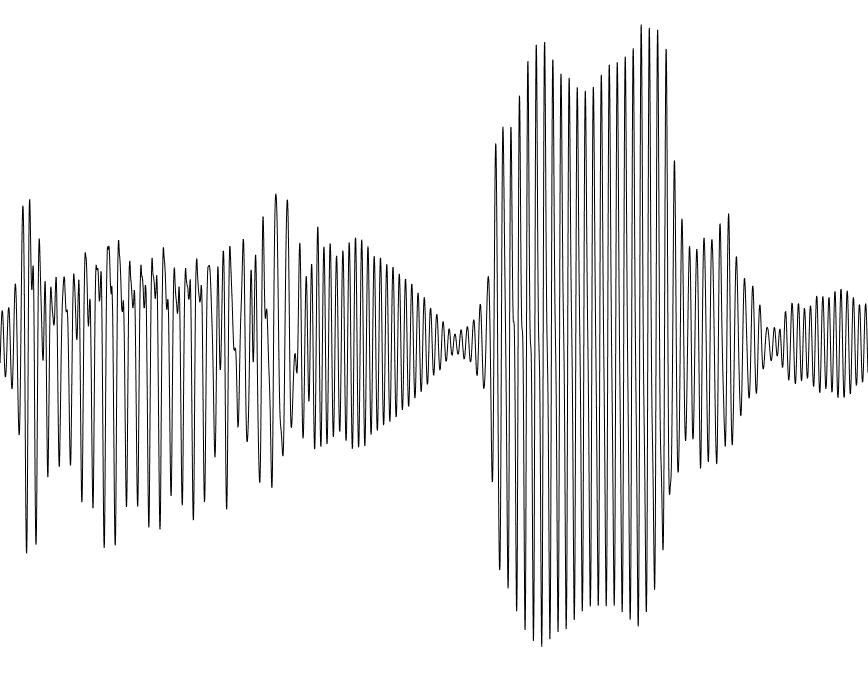};
\addplot [color=black]
  table[row sep=crcr]{%
0	0\\
};
\addlegendentry{data1}

\addplot [color=red, line width=1.0pt]
  table[row sep=crcr]{%
2.26757369614512e-05	0.029798082088847\\
0.00342403628117914	0.054152112534977\\
0.00682539682539683	0.114769327293418\\
0.0102267573696145	0.220746668370992\\
0.0136281179138322	0.353202769779807\\
0.0170294784580499	0.455300133888012\\
0.0204308390022676	0.480079005792838\\
0.0238321995464853	0.446975709203307\\
0.0272335600907029	0.419916983878958\\
0.0306349206349206	0.423279681320672\\
0.0340362811791383	0.448609184599018\\
0.037437641723356	0.465239911590214\\
0.0408390022675737	0.482212792408482\\
0.0442403628117914	0.493660053745428\\
0.0476417233560091	0.494781993718388\\
0.0510430839002268	0.476194705217933\\
0.0544444444444444	0.451038649869131\\
0.0578458049886621	0.429743424435457\\
0.0612471655328798	0.420851226377341\\
0.0646485260770975	0.423331255015534\\
0.0680498866213152	0.436656959702847\\
0.0714512471655329	0.464437918986397\\
0.0748526077097506	0.492173186978896\\
0.0782539682539683	0.505818584612896\\
0.0816553287981859	0.503994789246504\\
0.0850566893424036	0.495529815626709\\
0.0884580498866213	0.485496520708385\\
0.091859410430839	0.482595689880662\\
0.0952607709750567	0.490521544711676\\
0.0986621315192744	0.50792324625725\\
0.102063492063492	0.516376442789002\\
0.10546485260771	0.515367461879089\\
0.108866213151927	0.505321995790143\\
0.112267573696145	0.497547680376526\\
0.115668934240363	0.494532164717892\\
0.119070294784581	0.498671339739561\\
0.122471655328798	0.514171373856085\\
0.125873015873016	0.535089890306721\\
0.129274376417234	0.546204966521597\\
0.132675736961451	0.553876086376634\\
0.136077097505669	0.565236416160517\\
0.139478458049887	0.573223963640244\\
0.142879818594104	0.585326105742751\\
0.146281179138322	0.586780909735497\\
0.14968253968254	0.567102841216547\\
0.153083900226757	0.532828872800488\\
0.156485260770975	0.497573602200798\\
0.159886621315193	0.451572042386466\\
0.16328798185941	0.382998036130044\\
0.166689342403628	0.298491916583869\\
0.170090702947846	0.214000854801397\\
0.173492063492064	0.147862932541976\\
0.176893424036281	0.104095882264542\\
0.180294784580499	0.0796691221967953\\
0.183696145124717	0.0597026072760369\\
0.187097505668934	0.0403826989423859\\
0.190498866213152	0.0269682161807556\\
0.19390022675737	0.01889337918119\\
0.197301587301587	0.0179686954309907\\
0.200702947845805	0.0216043295994518\\
0.204104308390023	0.0254153774722887\\
0.20750566893424	0.0261484795331233\\
0.210907029478458	0.0239118883078384\\
0.214308390022676	0.0204413132178351\\
0.217709750566893	0.0190115828446991\\
0.221111111111111	0.0187704158469759\\
0.224512471655329	0.0190280595893263\\
0.227913832199546	0.0192055650146373\\
0.231315192743764	0.0190248362046527\\
0.234716553287982	0.0191395670542547\\
0.2381179138322	0.0190064678166392\\
0.241519274376417	0.0186827412266821\\
0.244920634920635	0.0185042381225707\\
0.248321995464853	0.0181475646478568\\
0.25172335600907	0.0181135654087443\\
0.255124716553288	0.0180774819919233\\
0.258526077097506	0.0175869950551021\\
0.261927437641723	0.0173201309214618\\
0.265328798185941	0.0184343011126262\\
0.268730158730159	0.0230825304273629\\
0.272131519274376	0.0384403441766211\\
0.275532879818594	0.0839732928146823\\
0.278934240362812	0.179824900687853\\
0.282335600907029	0.324118698518482\\
0.285736961451247	0.487341901254516\\
0.289138321995465	0.625509478627927\\
0.292539682539683	0.718904037305021\\
0.2959410430839	0.771514696693058\\
0.299342403628118	0.796950153610224\\
0.302743764172336	0.805654460672001\\
0.306145124716553	0.804353545277764\\
0.309546485260771	0.796164531816793\\
0.312947845804989	0.781453064672588\\
0.316349206349206	0.761575677209655\\
0.319750566893424	0.741060940552158\\
0.323151927437642	0.723329472007147\\
0.326553287981859	0.707992190597494\\
0.329954648526077	0.6936531096151\\
0.333356009070295	0.681955803313052\\
0.336757369614512	0.675422834738081\\
0.34015873015873	0.675220738218128\\
0.343560090702948	0.681730316036796\\
0.346961451247166	0.694077689324862\\
0.350362811791383	0.708676933867039\\
0.353764172335601	0.722072401741432\\
0.357165532879819	0.734245318659048\\
0.360566893424036	0.745708044373972\\
0.363968253968254	0.755262733147745\\
0.367369614512472	0.759851427997367\\
0.370770975056689	0.75379885600351\\
0.374172335600907	0.731719849606565\\
0.377573696145125	0.68878111863217\\
0.380975056689342	0.618053376596119\\
0.38437641723356	0.516924637066641\\
0.387777777777778	0.396524584324465\\
0.391179138321995	0.288710149811767\\
0.394580498866213	0.217598110220926\\
0.397981859410431	0.189649981853551\\
0.401383219954649	0.196831707272144\\
0.404784580498866	0.222397218204137\\
0.408185941043084	0.246572065403606\\
0.411587301587302	0.257560450260983\\
0.414988662131519	0.248949960021388\\
0.418390022675737	0.221872137399191\\
0.421791383219955	0.183525722764175\\
0.425192743764172	0.142881594552635\\
0.42859410430839	0.106756448532556\\
0.431995464852608	0.077882813223694\\
0.435396825396825	0.0571352628833536\\
0.438798185941043	0.0436463606213025\\
0.442199546485261	0.0360574609490767\\
0.445600907029478	0.0317796359317971\\
0.449002267573696	0.0288458800385005\\
0.452403628117914	0.0270920401314963\\
0.455804988662132	0.025945122961973\\
0.459206349206349	0.0243196579017167\\
0.462607709750567	0.0223468552335919\\
0.466009070294785	0.0213827896710085\\
0.469410430839002	0.0218711912747232\\
0.47281179138322	0.0223167879833588\\
0.476213151927438	0.0224396396054913\\
0.479614512471655	0.0233649656839265\\
0.483015873015873	0.024070100496458\\
0.486417233560091	0.0234055497981554\\
0.489818594104308	0.0216337555407308\\
0.493219954648526	0.0195197076764338\\
0.496621315192744	0.0182439923958056\\
};
\addlegendentry{data2}

\end{axis}
\end{tikzpicture}%
    	\label{fig:audio-app4}  
    \end{subfigure}
	\vspace*{-2em}
	\caption{Analysis of a recording of female speech (a), duration 0.5 seconds, sampled at 44.1 kHz, $n=22{,}050$. The three-component GP prior is overly simple given the true harmonic structure of the data, but the model is able to uncover high-, medium-, and low-frequency behaviour (b)-(d) along with their positive amplitude envelopes (shown in {\color{red}red}). Only the posterior means are shown. The posterior for the signal (not shown) is produced by multiplying the periodic components by their amplitudes and summing the three resulting signals (see \cref{sec:results} for more details regarding the model). The sub-components have been rescaled for visualisation purposes.}
	\label{fig:audio-app}
	\vspace*{-2em}
\end{figure}
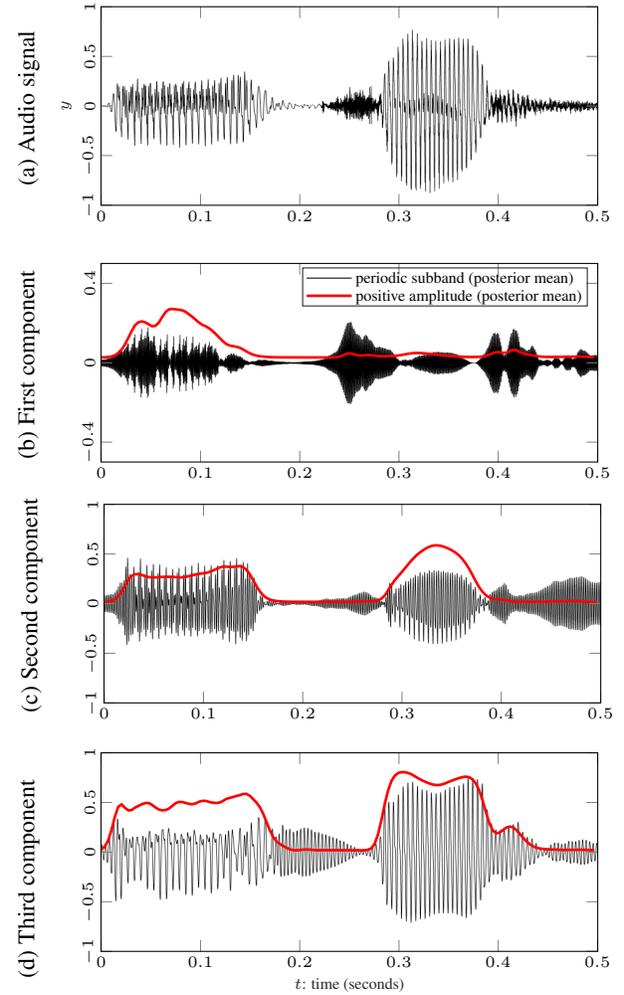

\onecolumn

\subsection{Motorcycle}

\end{document}